\documentclass{article} 
\usepackage{iclr2026_conference,times}
\usepackage{graphicx}
\usepackage{caption}

\usepackage{amsmath,amsfonts,bm}

\def\eqref#1{equation~\ref{#1}}

\def\1{\bm{1}}

\DeclareMathAlphabet{\mathsfit}{\encodingdefault}{\sfdefault}{m}{sl}
\SetMathAlphabet{\mathsfit}{bold}{\encodingdefault}{\sfdefault}{bx}{n}

\usepackage{wrapfig}
\usepackage{hyperref}
\usepackage{url}
\usepackage{booktabs} 
\usepackage{subcaption}
\usepackage{booktabs}         
\usepackage{siunitx}          
\usepackage{arydshln}         

\title{MathOPEval: A Fine-grained Benchmark  for Visual Operations of MLLMs in Mathematical Reasoning}

\author{
  Xiaoyuan Li\textsuperscript{1}\thanks{Work done when Xiaoyuan Li was intern at Alibaba Group.},  
  Moxin Li\textsuperscript{3}, 
  Wenjie Wang\textsuperscript{1}, 
  Rui Men\textsuperscript{2}, 
  Yichang Zhang\textsuperscript{2}, 
  Fuli Feng\textsuperscript{1},  \\
  \textbf{Dayiheng Liu}\textsuperscript{2} \\
  \textit{University of Science and Technology of China}\textsuperscript{1} 
  \textit{Alibaba Group}\textsuperscript{2} 
  \textit{National University of Singapore}\textsuperscript{3}
}

\iclrfinalcopy 
\begin{document}

\maketitle
\begin{abstract}
Recent progress in Multi-modal Large Language Models (MLLMs) has enabled step-by-step multi-modal mathematical reasoning by performing visual operations based on the textual instructions. 
A promising approach uses code as an intermediate representation to precisely express and manipulate the images in the reasoning steps. 
However, existing evaluations focus mainly on text-only reasoning outputs, leaving the MLLM's ability to perform accurate visual operations via code largely unexplored. This work takes a first step toward addressing that gap by evaluating MLLMs’ code-based capabilities in multi-modal mathematical reasoning.
Specifically, our framework focuses on two key evaluation aspects:  
(1) \textbf{Multi-modal Code Generation (MCG)} evaluates the model's ability to accurately understand and construct visualizations from scratch.  
(2) \textbf{Multi-modal Code Editing (MCE)} assesses the model's capacity for fine-grained operations, which include three types: \textbf{Deletion}, \textbf{Modification} and \textbf{Annotation}.
To evaluate the above tasks, we incorporate a dataset that covers the five most popular types of mathematical figures, including geometric diagrams, function plots, and three types of statistical charts, to provide a comprehensive and effective measurement of existing MLLMs. Our experimental evaluation involves nine mainstream MLLMs, and the results reveal that existing models still lag significantly behind human performance in performing fine-grained visual operations. 
\end{abstract}

\section{Introduction}

\begin{wrapfigure}{l}{0.5\linewidth}
    \centering 
    \setlength{\abovecaptionskip}{-0.10cm}
    \setlength{\belowcaptionskip}{0cm}
    
    \includegraphics[width=\linewidth]{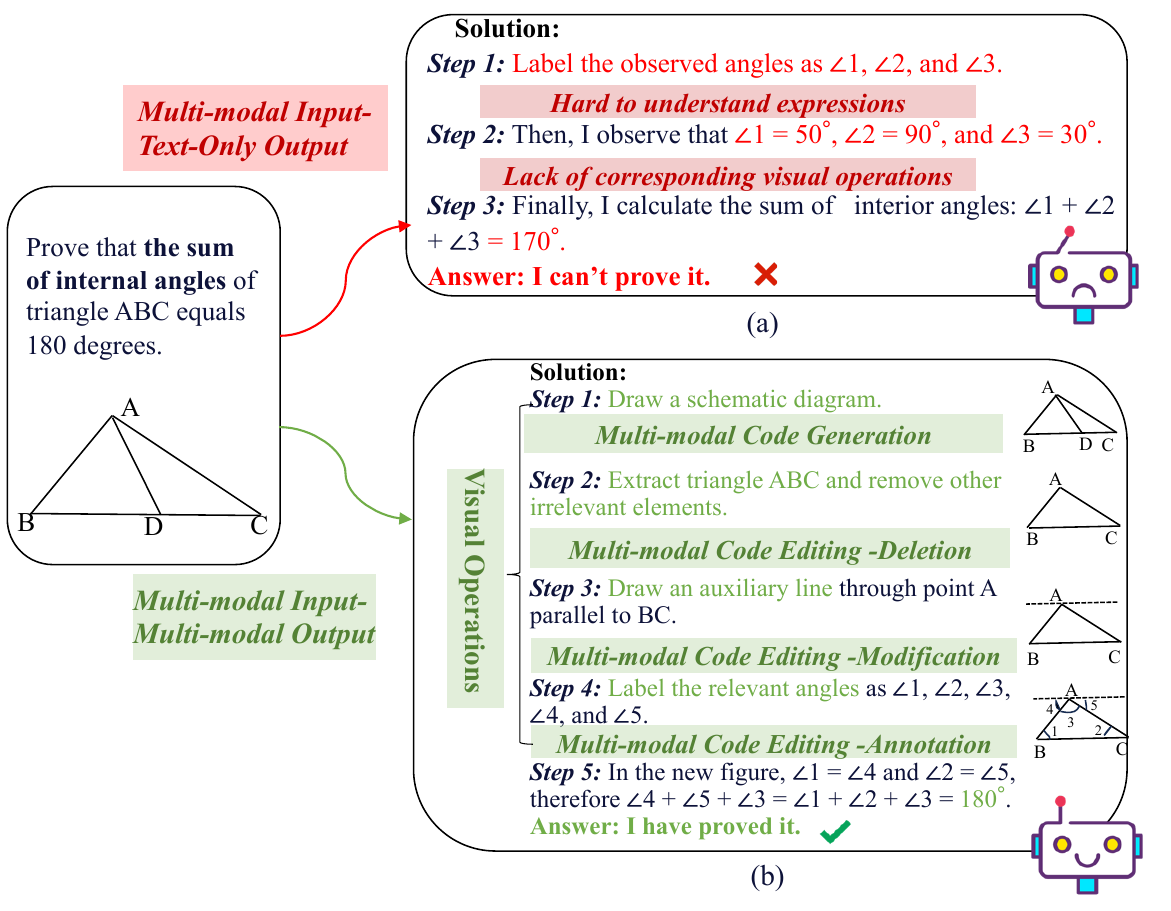} 
    
    \captionsetup{justification=raggedright} 
    \caption{Comparison of the paradigm of (a) multi-modal input, text-only output and (b) multi-modal input, multi-modal output with four types of visual operations.}
    \label{example}
\end{wrapfigure}
\vspace{-1em}
Recent years have witnessed remarkable progress in visual mathematical reasoning by Multi-modal Large Language Models (MLLMs) \citep{DBLP:journals/corr/abs-2412-15115,DBLP:journals/corr/abs-2410-21276,DBLP:journals/corr/abs-2403-05530}. Given an image and a textual problem description, MLLMs generate step-by-step reasoning toward the solution. These reasoning steps often include visual operations such as annotating angles, drawing auxiliary lines, and identifying key elements (Figure~\ref{example}). Compared to text-only mathematical reasoning, this task demands stronger multi-modal capabilities, particularly in aligning textual and visual information,  planning and conducting visual operations to reach the correct answer. 

Recently, using code as an intermediate representation for multi-modal mathematical reasoning has emerged as a promising approach, leading to substantial performance gains \citep{DBLP:conf/nips/HuSFROZSK24,DBLP:journals/corr/abs-2501-05452}. 
In this paradigm, images are translated into corresponding Python or \LaTeX~codes that can accurately reconstruct them. Visual operations are then represented as code edits. Code serves as a precise and structured textual representation of images, aligning well with language-based reasoning and reducing the ambiguity often present in direct visual operations.
However, recent evaluations of multi-modal mathematical reasoning have primarily focused on text-only outputs \citep{DBLP:conf/nips/HuSFROZSK24, DBLP:journals/corr/abs-2412-12932,DBLP:journals/corr/abs-2501-05452}. It remains unclear how effectively current MLLMs can generate intermediate code that reflects accurate visual operations—an essential step toward precise, interpretable multi-modal mathematical reasoning (Figure~\ref{example}). 

\begin{figure*}[t]   
\centering
\setlength{\abovecaptionskip}{-0.10cm}
\setlength{\belowcaptionskip}{0cm}
\includegraphics[width=\linewidth,scale=1.0]{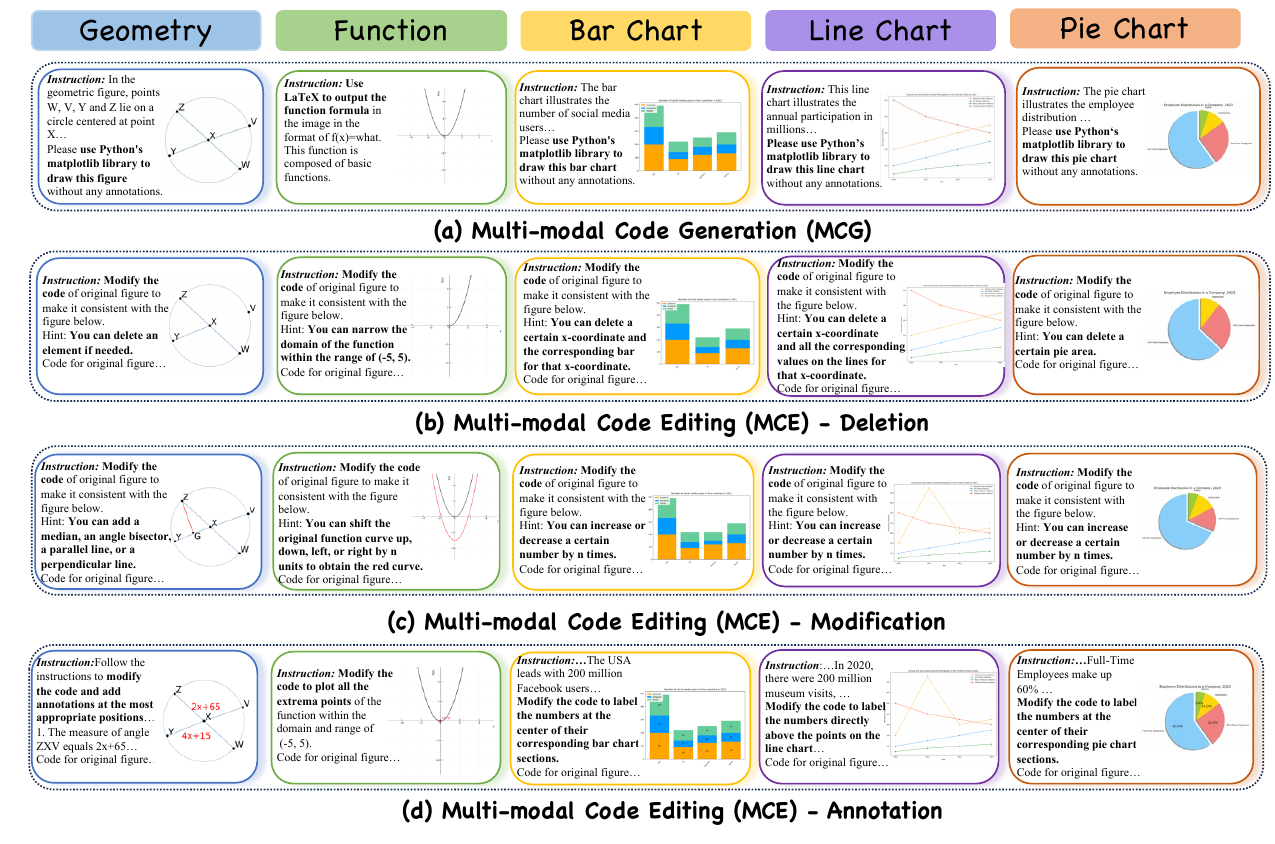}
\caption{Illustration of initial dataset for four visual operations across five visualization types. The dataset includes evaluation instructions, code, and images constructed for different tasks and visualization types. Please refer to Section \ref{dataset} for details. }
\label{framework}
\vspace{-2em}
\vspace{-5pt}
\end{figure*}

Therefore, in this work, we take a first step toward evaluating multi-modal mathematical reasoning by focusing on MLLMs' code-related capabilities in performing visual operations. 
We focus on two primary aspects: \textbf{(1) Multi-modal Code Generation} and \textbf{(2) Multi-modal Code Editing}. Multi-modal Code Generation evaluates models' ability to transform the image input into the corresponding construction code. Multi-modal Code Editing encompasses three common types of visual operations: \textbf{1) Deletion} tests models' capability to identify and remove distracting elements, streamlining figures to highlight key information; \textbf{2) Modification}: evaluates models' proficiency in updating images through adding auxiliary lines or altering visual elements; \textbf{3) Annotation} assesses models' ability to add numerical values, symbols, and other markings at appropriate locations, enhancing visual information expression through clear annotations.

Move one step further, we manually curate a high-quality dataset. Considering the diversity of mathematical visual reasoning tasks, we carefully select five representative types of graphics: \textbf{(1) geometric figures}, \textbf{(2) function graphs} , \textbf{(3) bar charts},  \textbf{(4) line graphs}, and \textbf{(5) pie charts}. For each sample, we conduct detailed manual annotations to construct initial dataset as shown in Figure \ref{framework}, including: (1) original question image; (2) images after four visual operations; (3) codes and instructions for implementing these visual operations. To ensure comprehensive and reliable evaluation, we transform the initial dataset into two question types: 10,293 multiple-choice questions to test models' selection ability of visual operations, and 7,552 open-ended questions to assess models' ability to generate visual operations. For free-generation questions, we also establish a evaluation strategy based on Chain-of-Thought (CoT) to conduct fine-grained assessment across multiple dimensions, including content accuracy, position consistency, code correctness and completeness.

We conduct comprehensive experiments on nine mainstream closed-source and open-source MLLMs using diverse prompting strategies. Our experimental results reveal these \textbf{key findings}: (1) Existing MLLMs still show significant performance gaps compared to human-level capability on our visual operation evaluation suite, indicating that our benchmark poses substantial challenges to current MLLMs; (2) Among different types of visual operations, visual modification proves to be the most challenging, while among different types of diagrams, function plots present the highest level of difficulty in processing; (3) Increasing the number of shots can moderately improve models' understanding and execution capabilities in visual operations.

Our main contributions can be summarized as follows: 
\begin{itemize}
\item To our knowledge, we are the first to propose a fine-grained evaluation benchmark for visual operations, providing new perspectives and tools for in-depth assessment of MLLMs' visual understanding capabilities. 
\item We construct a carefully annotated large-scale visual operation evaluation dataset covering various mathematical graphic types and visual operation types.
\item Through extensive experiments, we reveal the capability boundary and improvement potential of MLLMs in visual operations, providing important references for future research.
\end{itemize}

\section{Task Formulation}

This work evaluates the multi-modal visual operation capabilities of MLLMs in the form of multi-modal coding, focusing on two key aspects: multi-modal code generation and multi-modal code editing. These aspects reflect common patterns in multi-modal mathematical reasoning, where models must translate image (e.g., function graphs or charts) into code and make precise, coordinated edits to code based on image edits. 

\begin{itemize}
    \item \textbf{Multi-modal Code Generation (MCG)}: Given an image $V_{in}$, MLLM generate the corresponding code $C_{in}$ that can precisely construct such image. The code can either be Python Matplotlib, or \LaTeX. Formally, denoting the generation instruction as $I_{g}$. 
    \begin{align}
        C_{in} = MLLM(V_{in}, I_g) \label{eq:mcg}
    \end{align}
    \item \textbf{Multi-modal Code Editing (MCE)}: MLLM is given an initial image $V_{in}$, its code $C_{in}$, and an edited image $V_{out}$. MLLM is expected to generate the code for $V_{out}$, denoted as $C_{out}$ by editing the initial code $C_{in}$. Formally, denoting the editing instruction as $I_{e}$. 
    \begin{align}
        C_{out} = MLLM(V_{in}, V_{out}, C_{in}, I_e) \label{eq:mce}
    \end{align}
\end{itemize}

\begin{itemize}
    \item \textbf{MCE (Deletion)}: $V_{out}$ is obtained by removing certain elements from $V_{in}$—for example, deleting a pie area in the pie chart, or deleting a bar in the bar chart. 
    \item \textbf{MCE (Modification)}: $V_{out}$ is derived from $V_{in}$ by modifying specific elements, such as adding an auxiliary line in the geometric plot, or adjusting bar height in the bar chart.
    \item \textbf{MCE (Annotation)}: $V_{out}$ is obtained by locating and annotating the value for certain elements in $V_{in}$. 
    Different from the other tasks, We deliberately omit $V_{out}$ from the input to assess the model's ability to place value annotations in appropriate positions. This approach allows us to assess more deeply how well the model understands proper label placement.  
\end{itemize}

For the \textbf{MCE}, we scrutinize multi-modal mathematical reasoning examples and thoroughly define three types of editing operations. In total, we define four evaluation tasks.


\section{Dataset Construction}
\label{dataset}
A significant challenge in implementing these evaluation tasks is the lack of dataset to perform such evaluation. 
To address this limitation, we manually construct a dataset \texttt{MathOPEval} (Visual \textbf{Math}ematical \textbf{OP}eration \textbf{Eval}uation) that meets our specific evaluation requirements. The construction process involves two stages: \textbf{(1) creating an initial dataset} $D_{\text{init}}$ containing codes, images, and instructions for the four task types and \textbf{(2) transforming this dataset into two formats} - free-form generation $D_{\text{gen}}$ and multiple-choice questions $D_{\text{mc}}$. 

\subsection{Initial Dataset Construction}

Our initial dataset $D_{\text{init}}$ consists of images $V$, their corresponding construction codes $C$, and textual instructions $I$. For each image, we construct three image variations along with their codes and instructions corresponding to the four evaluation tasks. Specifically, 

\begin{itemize}
\item Original state ($V_{\text{orig}}$, $C_{\text{orig}}$, $I_{\text{orig}}$): The original image, its code, and a textual description. 
\item Deleted state ($V_{\text{del}}$, $C_{\text{del}}$, $I_{\text{del}}$): The image with elements deleted from $V_{\text{orig}}$, its code, and a hint on the possible deletion operations. 
\item Modified state ($V_{\text{mod}}$, $C_{\text{mod}}$, $I_{\text{mod}}$): The image with elements modified from $V_{\text{orig}}$, its code, and a hint on the possible modifications. 
\item Annotated state ($V_{\text{ann}}$, $C_{\text{ann}}$, $I_{\text{ann}}$): The annotated image, its code and the corresponding annotation instruction containing the elements and values for annotation, such as angle measures, coordinates, or numerical values.
\end{itemize}

To ensure the diversity of the image types, we consider three sources, five types of mathematical visualizations: \textbf{geometric figures, function plots, statistical charts including bar charts, line graphs, and pie charts}.
For each type of mathematical visualization, we consider the diversity of their context domains. We delineate our source of data and construction details as follow.

\begin{itemize}
   \item \textbf{Geometric Figures:}  Starting with geometric problems from the Geometry3K dataset \citep{DBLP:conf/acl/LuGJQHLZ20}, we first obtain $V_{\text{orig}}$ and manually annotate the image descriptions ($I_{\text{orig}}$) and code $C_{\text{orig}}$. The deletion state involves removing specified line segments to simplify the visualization, while the modification state incorporates auxiliary geometric elements such as medians, perpendicular lines, angle bisectors, and parallel lines to facilitate problem-solving. For the annotation state, we enhance figures with point labels and line measurements from problem descriptions.

    \item \textbf{Function Plots:} We develop a standardized approach leveraging common high school-level functions collected from the internet\footnote{https://homework.study.com}\footnote{https://mathspace.co/us}. We set the domain of the functions as $[-5,5]$, and generate initial plots and instructions ($V_{\text{orig}}$, $C_{\text{orig}}, I_{\text{orig}}$). For the deletion stage, we delete part of the domain. The modification stage implements various function transformations and the annotation phase focuses on marking critical points, including x-axis and y-axis intersections and extreme points.

    \item \textbf{Statistical Charts:} We consider three types of charts: bar charts, line graphs, and pie charts. 
    We collect the original state ($V_{\text{orig}}$, $C_{\text{orig}}$, $I_{\text{orig}}$) from the ChartX dataset \citep{DBLP:journals/corr/abs-2402-12185}. 
    In the deletion state, we remove selected elements from the visualizations: bars from bar charts, points at specific x-coordinates from line graphs, and areas from pie charts. 
    In the modification state, we adjust specific visual properties—such as changing bar heights in bar charts, altering y-coordinates of selected points in line graphs, and modifying sector proportions in pie charts. 
    In the annotation state, we add numerical labels to represent key values: bar heights in bar charts, y-coordinates of points in line graphs, and sector percentages in pie charts. 

\end{itemize}

\subsection{Dataset Format Conversion}
Following the definition in Eq~\ref{eq:mcg} and Eq~\ref{eq:mce}, we transform the initial dataset $D_{\text{init}}$ into evaluation formats. 
Specifically, for free-form generation dataset ($D_\text{gen}$), 
\begin{figure}[t] 

    \begin{minipage}[c]{0.48\linewidth} 
        \centering
        \includegraphics[width=0.95\linewidth]{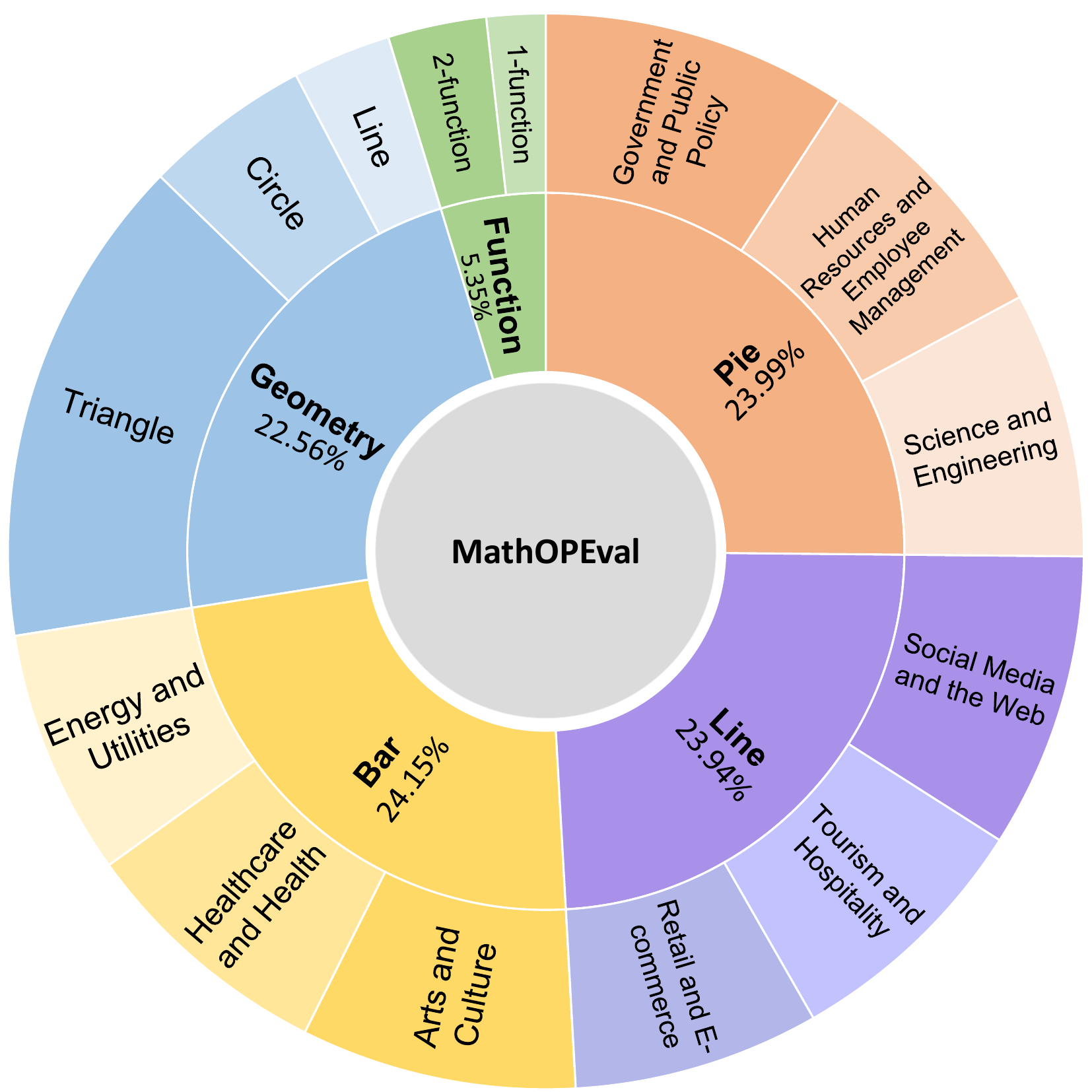} 
        \captionsetup{justification=raggedright, type=figure}
        \caption{Distribution of five visualization types and their major context domains.}
        \label{fig:sta_dist}
    \end{minipage}%
    \hfill
    \begin{minipage}[c]{0.48\linewidth}
        \centering
        \small 
        \setlength{\tabcolsep}{4pt}
        \begin{tabular}{ll}
            \toprule
            \textbf{Statistic}        & \textbf{Number} \\ \midrule
            Total samples             & 1,888            \\
            \small{- Geometric figures} & \small{426}           \\
            \small{- Function plots}   & \small{101}          \\ 
            \small{- Bar charts}   & \small{456}          \\ 
            \small{- Line graphs}   & \small{452}          \\ 
            \small{- Pie charts}   & \small{453}          \\ \midrule
            Total questions           & 17,845           \\ 
            Total images             & 7,552            \\ \midrule
            Multiple-choice questions & 10,293           \\
            \small{- Average question length} & \small{89.52}           \\
            \small{- Average choice length}   & \small{132.12}          \\ \midrule
            Free-form questions       & 7,552            \\
            \small{- Average question length} & \small{868.64}          \\
            \small{- Average answer length}   & \small{783.77}          \\ \bottomrule
        \end{tabular}
        \captionsetup{justification=raggedright, type=table}
        \caption{Basic statistics of MathOPEval.}
        \label{tab:sta_stats}
    \end{minipage}

\end{figure}

\begin{itemize}
    \item \textbf{MCG:} $C_\text{orig} = MLLM(V_\text{orig}, I_\text{orig})$
    \item \textbf{MCE (Deletion):} $C_\text{del} = MLLM(V_\text{orig}, V_\text{del}, C_\text{orig}, I_\text{del})$
    \item \textbf{MCE (Modification):} $C_\text{mod} = MLLM(V_\text{orig}, V_\text{mod}, C_\text{orig}, I_\text{mod})$
    \item \textbf{MCE (Annotation):} $C_\text{ann} = MLLM(V_\text{orig}, C_\text{orig}, I_\text{ann})$
\end{itemize}

For multiple-choice dataset ($D_\text{mc}$), we further augment each task with three carefully designed incorrect options while maintaining similar format and length.

\subsection{Dataset Statistics}
The statistics of \texttt{MathOPEval} is shown in Table~\ref{tab:sta_stats}. 
In the initial dataset $D_{\text{init}}$, each sample contains four sets of task-specific codes, images, and instructions. After format conversion, we obtain 7,552 questions in the generation dataset $D_{\text{gen}}$ and 10,293 questions in the multiple-choice dataset $D_{\text{mc}}$, resulting in a total of 17,845 questions. 
Figure~\ref{fig:sta_dist} shows the statistics on the proportion of different types of visualization input and lists their major context domains. We can observe that the dataset covers a diverse range of domains, including ``Arts and Culture'', ``Social Media and the Web'', demonstrating its broad applicability across different contexts.

\section{Chain-of-Thought Evaluation Strategy} \label{sec:evaluation_strategy}

After constructing the evaluation dataset, we aim to develop an effective and efficient strategy for assessing the correctness of generated code outputs. 
While evaluation is straightforward in the multiple-choice format—by extracting answer choices and computing accuracy—it becomes more challenging in free-form generation settings. 
Given the high cost of human evaluation, we leverage state-of-the-art LLMs with strong language and coding capabilities. Specifically, we propose an automated Chain-of-Thought (CoT) evaluation strategy, utilizing the DeepSeek-V3~\citep{deepseekv3} model for reliable code evaluation. Our evaluation process comprises two key steps: \textbf{key element extraction} and \textbf{analytical scoring}. 

\noindent
\textbf{Key Element Extraction} 
To ensure precise and efficient evaluation, we instruct the evaluator model to first extract and compare key elements from both the generated and reference codes, with the specific elements varying by task type. This targeted comparison enables more reliable evaluation of code correctness and task-specific MLLM coding capabilities.
\begin{itemize}
    \item In \textbf{MCG}, the extracted elements vary by visualization type: mathematical expressions for function graphs, geometric components like points, lines, and shapes for geometric figures, and data components for statistical charts. For bar charts, we focus on bar heights, category labels, bar quantity and arrangement, and structural organization. Line graphs require analysis of data points, connection styles, and point density. Pie charts are evaluated based on slice proportions, category labels, and slice ordering. We extract these elements from the generated and the reference $C_{out}$ to examine their identity. 
    \item In \textbf{MCE}, we analyze the differences between the generated output and the reference output by comparing both versions of $C_{out}$ against the original input $C_{in}$. 
\end{itemize}

\noindent
\textbf{Analytical Scoring} 
We design a comprehensive five-level scoring rule (1–2, 3–4, 5–6, 7–8, 9–10) that evaluates outputs across multiple dimensions: content accuracy, positional consistency, code correctness, and completeness. For each visualization type and task, we define detailed evaluation criteria corresponding to each score range, as outlined in the Appendix.


\noindent
\textbf{Human Evaluation} To assess the reliability of our automated evaluation, we randomly sampled 100 model-generated scores and conducted human evaluations. Results show that 92\% of the scores were considered reasonable by human annotators, indicating strong alignment between automated and manual judgments. Full details of the human evaluation criteria are provided in the Appendix. 
\section{Experiments}

\subsection{Experimental Setup}

We conduct evaluations on a diverse set of models, including proprietary models GPT-4o \citep{DBLP:journals/corr/abs-2410-21276}, Qwen-VL-Max \citep{DBLP:journals/corr/abs-2412-15115}, and open-source models Qwen2.5-VL \citep{DBLP:journals/corr/abs-2412-15115}, Gemma3 \citep{DBLP:journals/corr/abs-2403-05530}, LLaVA-NeXT \citep{DBLP:conf/cvpr/LiuLLL24} Series, and specialized reasoning model QVQ-72B \citep{DBLP:journals/corr/abs-2412-15115}. To minimize the randomness, we set the temperature parameter to 0. For multiple-choice tasks, we investigate various prompting strategies: \textbf{(1) Direct} instructs models to output answers directly; \textbf{(2) CoT} \citep{DBLP:conf/nips/Wei0SBIXCLZ22} elicits step-by-step reasoning paths; \textbf{(3) Descriptive Chain-of-Thought (DCoT)} \citep{DBLP:journals/corr/abs-2311-09193} requires models to generate descriptions before answering; and \textbf{(4) Visualization-of-Thought (VoT)} \citep{DBLP:conf/nips/WuMZ000W24} prompts models to imagine reasoning paths with the instruction ``Visualize the state after each reasoning step.'' Due to resource constraints, we evaluate Direct prompting for free-form generation tasks.

\subsection{Main Results}

Tables \ref{tab:main experiment 1.} and \ref{tab:main experiment 2.} present main results for multiple-choice questions and free-form generation tasks. Our analysis of these results reveals several key findings:

\noindent
\textbf{Model Performance:} A substantial performance gap exists between models and human baseline across all tasks. Visual operations that humans find intuitive pose significant challenges for models, with an average performance gap of 52.39\% and peaking at 67.13\% for geometric deletion tasks. Among the evaluated models, Qwen2.5-VL-72B-Instruct achieves the highest accuracy of 68.02\% in multiple-choice questions, while GPT-4o demonstrates superior performance in free-form generation with an average score of 6.08. At the other end of the spectrum, Gemma3-4B-IT consistently shows the lowest performance, scoring 21.00\% and 3.30 in multiple-choice and free-form generation respectively. The results demonstrate that generally, visual operation performance improves with model size, aligning with established scaling laws in language models \citep{DBLP:journals/corr/abs-2001-08361}.

\begin{table}[t]
\centering
\renewcommand{\arraystretch}{1.3}
\setlength{\abovecaptionskip}{0.05cm}
\setlength{\belowcaptionskip}{0cm}
\scalebox{0.49}{
\begin{tabular}{llllllllllllllllllllll}
\hline
\multicolumn{1}{l|}{}       & \multicolumn{5}{c|}{\textbf{MCG}}                       & \multicolumn{5}{c|}{\textbf{MCE (Deletion)}}                      & \multicolumn{5}{c|}{\textbf{MCE (Modification)}}                            & \multicolumn{5}{c|}{\textbf{MCE (Annotation)}}                        &       \\ \hline
\multicolumn{1}{l|}{}       & Geo   & Func  & Bar   & Line  & \multicolumn{1}{l|}{Pie}   & Geo   & Func   & Bar   & Line  & \multicolumn{1}{l|}{Pie}   & Geo    & Func   & Bar    & Line   & \multicolumn{1}{l|}{Pie}    & Geo   & Func  & Bar   & Line   & \multicolumn{1}{l|}{Pie}   & AVG   \\ \hline
\multicolumn{1}{l|}{Human} & 93.33 & 91.67 & 98.33 & 96.67 & \multicolumn{1}{l|}{96.67} & 100.00 & 100.00 & 100.00 & 100.00 & \multicolumn{1}{l|}{100.00} & 96.67 & 98.33 & 98.33 & 100.00 & \multicolumn{1}{l|}{98.33} & 98.33 & 100.00 & 96.67 & 98.33 & \multicolumn{1}{l|}{96.67} & 97.92 \\ \hline
\multicolumn{22}{c}{\textit{GPT-4o}}                                                                                                                                                                                                                                                                 \\
\cdashline{1-22} 
\multicolumn{1}{l|}{Direct} & 71.36 & \underline{\textbf{51.49}} & 50.22 & 73.89 & \multicolumn{1}{l|}{40.18} & 33.57 & \underline{44.55} & 67.76 & 48.01 & \multicolumn{1}{l|}{80.13} & 55.63 & 43.56 & 56.58 & 62.61 & \multicolumn{1}{l|}{49.89} & 44.63 & 73.79 & 45.39 & 37.61 & \multicolumn{1}{l|}{64.24} & 54.76 \\
\multicolumn{1}{l|}{CoT} & 68.08 & \underline{\textbf{51.49}} & \underline{64.25} & 80.31 & \multicolumn{1}{l|}{49.89} & 49.30 & \underline{44.55} & \underline{97.81} & 68.81 & \multicolumn{1}{l|}{96.47} & \underline{77.93} & 45.54 & \underline{\textbf{71.49}} & 76.99 & \multicolumn{1}{l|}{46.14} & 39.91 & 71.05 & 55.04 & 53.10 & \multicolumn{1}{l|}{78.81} & 64.35 \\
\multicolumn{1}{l|}{DCoT} & \underline{78.17} & 50.50 & 58.77 & \underline{83.85} & \multicolumn{1}{l|}{\underline{50.99}} & 48.59 & 22.77 & \underline{97.81} & 67.26 & \multicolumn{1}{l|}{96.91} & 77.00 & 48.51 & 69.30 & 75.88 & \multicolumn{1}{l|}{47.68} & \underline{54.59} & \underline{75.15} & \underline{69.74} & \underline{66.59} & \multicolumn{1}{l|}{80.35} & \underline{66.02} \\
\multicolumn{1}{l|}{VCoT} & 61.74 & 45.54 & 54.82 & 80.53 & \multicolumn{1}{l|}{47.68} & \underline{51.88} & 39.60 & 97.37 & \underline{70.35} & \multicolumn{1}{l|}{\underline{98.45}} & 74.88 & \underline{51.49} & 69.74 & \underline{79.65} & \multicolumn{1}{l|}{\underline{\textbf{50.33}}} & 41.59 & 69.83 & 55.48 & 52.88 & \multicolumn{1}{l|}{\underline{80.79}} & 63.73 \\ \hline
\multicolumn{22}{c}{\textit{Qwen-VL-Max}}                                                                                                                                                                                                                                                               \\\cdashline{1-22} 

\multicolumn{1}{l|}{Direct} & \underline{84.51} & 43.56 & 41.45 & 58.85 & \multicolumn{1}{l|}{43.93} & \underline{\textbf{52.82}} & 59.41 & 91.89 & 69.91 & \multicolumn{1}{l|}{98.45} & 68.54 & 44.55 & 67.11 & 60.62 & \multicolumn{1}{l|}{\underline{54.30}} & 50.06 & \underline{70.43} & 51.75 & 50.22 & \multicolumn{1}{l|}{71.52} & 61.69 \\
\multicolumn{1}{l|}{CoT} & 79.11 & \underline{46.53} & 84.43 & \underline{90.93} & \multicolumn{1}{l|}{\underline{49.67}} & 45.77 & \underline{68.32} & \underline{97.15} & 72.35 & \multicolumn{1}{l|}{\underline{98.90}} & \underline{81.92} & \underline{47.52} & 76.54 & \underline{78.98} & \multicolumn{1}{l|}{48.79} & 48.71 & 67.62 & 61.62 & 47.57 & \multicolumn{1}{l|}{62.25} & 67.73 \\
\multicolumn{1}{l|}{DCoT} & 73.00 & 43.56 & 69.52 & 71.02 & \multicolumn{1}{l|}{44.37} & 51.64 & 63.37 & 92.11 & 52.65 & \multicolumn{1}{l|}{98.23} & 77.70 & 45.54 & 75.44 & 77.21 & \multicolumn{1}{l|}{45.92} & \underline{51.55} & 70.17 & \underline{73.25} & \underline{66.37} & \multicolumn{1}{l|}{\underline{71.96}} & 65.73 \\
\multicolumn{1}{l|}{VCoT} & 77.46 & 45.54 & \underline{85.96} & 86.06 & \multicolumn{1}{l|}{48.34} & 50.94 & 63.37 & 96.71 & \underline{72.79} & \multicolumn{1}{l|}{\underline{98.90}} & 81.46 & 46.53 & \underline{77.63} & 78.32 & \multicolumn{1}{l|}{51.43} & 48.06 & 66.06 & 64.47 & 50.66 & \multicolumn{1}{l|}{66.67} & \underline{67.87} \\
\hline

\multicolumn{22}{c}{\textit{Qwen2.5-VL-3B-Instruct}}                                                                                                                                                                                                                                                    \\
\cdashline{1-22} 
\multicolumn{1}{l|}{Direct} & 44.55 & \underline{40.59} & \underline{37.50} & \underline{57.08} & \multicolumn{1}{l|}{\underline{52.54}} & 30.28 & 26.73 & \underline{75.00} & \underline{24.34} & \multicolumn{1}{l|}{\underline{96.47}} & 46.95 & \underline{42.57} & \underline{37.50} & \underline{58.85} & \multicolumn{1}{l|}{19.65} & \underline{52.39} & 68.33 & 69.74 & 63.05 & \multicolumn{1}{l|}{\underline{82.34}} & \underline{51.32} \\
\multicolumn{1}{l|}{CoT} & \underline{47.52} & 32.67 & 30.04 & 45.35 & \multicolumn{1}{l|}{41.72} & \underline{38.21} & 22.77 & 37.06 & 14.38 & \multicolumn{1}{l|}{61.59} & 42.02 & 31.68 & 34.65 & 47.35 & \multicolumn{1}{l|}{\underline{26.49}} & 18.76 & 67.39 & 63.60 & 54.42 & \multicolumn{1}{l|}{66.89} & 41.23 \\
\multicolumn{1}{l|}{DCoT} & 41.58 & 39.60 & 32.68 & 38.27 & \multicolumn{1}{l|}{45.03} & 31.46 & \underline{31.68} & 69.30 & 21.68 & \multicolumn{1}{l|}{92.05} & \underline{49.06} & 41.58 & 30.70 & 50.22 & \multicolumn{1}{l|}{19.87} & 41.91 & 61.22 & \underline{74.78} & \underline{69.69} & \multicolumn{1}{l|}{73.95} & 47.82 \\
\multicolumn{1}{l|}{VCoT} & 42.57 & 30.69 & 33.77 & 46.46 & \multicolumn{1}{l|}{47.46} & 31.92 & 23.76 & 70.39 & 21.02 & \multicolumn{1}{l|}{93.16} & 47.42 & 37.62 & 31.36 & 52.21 & \multicolumn{1}{l|}{21.85} & 20.31 & \underline{73.00} & 72.37 & 64.38 & \multicolumn{1}{l|}{78.59} & 47.02 \\
\hline
\multicolumn{22}{c}{\textit{Qwen2.5-VL-72B-Instruct}}                                                                                                                                                                                                                                                   \\
\cdashline{1-22} 
\multicolumn{1}{l|}{Direct} & \underline{\textbf{91.55}} & 44.55 & 83.77 & \underline{\textbf{95.13}} & \multicolumn{1}{l|}{\underline{\textbf{71.96}}} & \underline{41.55} & 56.44 & \underline{\textbf{99.56}} & \underline{\textbf{75.22}} & \multicolumn{1}{l|}{\underline{\textbf{99.78}}} & 77.93 & \underline{\textbf{52.48}} & \underline{63.82} & 72.79 & \multicolumn{1}{l|}{\underline{35.98}} & \underline{\textbf{72.70}} & \underline{\textbf{79.62}} & \underline{\textbf{92.11}} & \underline{\textbf{91.15}} & \multicolumn{1}{l|}{\underline{\textbf{97.79}}} & \underline{\textbf{74.79}} \\
\multicolumn{1}{l|}{CoT} & 77.70 & \underline{47.52} & 87.94 & 93.58 & \multicolumn{1}{l|}{24.06} & 32.63 & 50.50 & 96.93 & 41.37 & \multicolumn{1}{l|}{98.90} & \underline{\textbf{82.16}} & 48.51 & 62.94 & 79.20 & \multicolumn{1}{l|}{26.05} & 50.06 & 68.68 & 77.63 & 54.65 & \multicolumn{1}{l|}{85.87} & 64.34 \\
\multicolumn{1}{l|}{DCoT} & 85.92 & 41.58 & 83.77 & 94.47 & \multicolumn{1}{l|}{25.61} & 32.63 & \underline{\textbf{59.41}} & 92.98 & 44.47 & \multicolumn{1}{l|}{99.56} & 81.22 & 46.53 & 58.11 & \underline{\textbf{79.87}} & \multicolumn{1}{l|}{35.54} & 54.40 & 75.00 & 78.95 & 63.50 & \multicolumn{1}{l|}{90.95} & 66.22 \\
\multicolumn{1}{l|}{VCoT} & 78.40 & 42.57 & \underline{\textbf{88.82}} & 93.81 & \multicolumn{1}{l|}{61.59} & 33.33 & 53.47 & 96.71 & 53.76 & \multicolumn{1}{l|}{98.90} & 81.69 & 46.53 & 59.87 & 78.98 & \multicolumn{1}{l|}{27.59} & 48.19 & 70.52 & 78.07 & 54.42 & \multicolumn{1}{l|}{86.75} & 66.70 \\
\hline
\multicolumn{22}{c}{\textit{Gemma3-4B-IT}}                                                                                                                                                                                                                                                              \\

\cdashline{1-22} 

\multicolumn{1}{l|}{Direct} & \underline{32.86} & \underline{26.73} & \underline{30.48} & \underline{24.78} & \multicolumn{1}{l|}{\underline{28.48}} & \underline{34.04} & 14.85 & 25.66 & 20.35 & \multicolumn{1}{l|}{\underline{37.53}} & \underline{31.22} & 24.75 & 33.77 & 35.40 & \multicolumn{1}{l|}{33.55} & \underline{29.43} & 9.89 & \underline{33.11} & \underline{29.42} & \multicolumn{1}{l|}{\underline{30.24}} & \underline{28.33} \\
\multicolumn{1}{l|}{CoT} & 10.09 & 21.78 & 26.75 & 24.56 & \multicolumn{1}{l|}{23.84} & 25.59 & \underline{17.82} & \underline{37.94} & 19.69 & \multicolumn{1}{l|}{23.84} & 28.40 & \underline{31.68} & \underline{41.89} & \underline{44.03} & \multicolumn{1}{l|}{\underline{41.28}} & 9.77 & \underline{23.07} & 25.88 & 14.60 & \multicolumn{1}{l|}{29.14} & 26.08 \\
\multicolumn{1}{l|}{DCoT} & 24.65 & 5.94 & 17.76 & 6.42 & \multicolumn{1}{l|}{20.53} & 12.21 & 0.99 & 20.39 & 0.88 & \multicolumn{1}{l|}{9.71} & 10.80 & 13.86 & 7.24 & 6.86 & \multicolumn{1}{l|}{0.22} & 0.19 & 2.83 & 0.88 & 1.11 & \multicolumn{1}{l|}{0.44} & 8.20 \\
\multicolumn{1}{l|}{VCoT} & 10.09 & 18.81 & 18.86 & 9.29 & \multicolumn{1}{l|}{22.52} & 14.79 & 9.90 & 8.55 & \underline{21.90} & \multicolumn{1}{l|}{22.30} & 23.71 & 30.69 & 38.38 & 34.96 & \multicolumn{1}{l|}{32.45} & 10.03 & 19.70 & 28.29 & 24.34 & \multicolumn{1}{l|}{28.70} & 21.41 \\
\hline
\multicolumn{22}{c}{\textit{Gemma3-27B-IT}}                                                                                                                                                                                                                                                             \\
\cdashline{1-22} 
\multicolumn{1}{l|}{Direct} & \underline{44.84} & 17.82 & 21.05 & 26.55 & \multicolumn{1}{l|}{34.00} & \underline{35.45} & 24.75 & \underline{22.15} & \underline{33.19} & \multicolumn{1}{l|}{25.39} & 29.58 & \underline{22.77} & 28.29 & 24.56 & \multicolumn{1}{l|}{43.71} & \underline{24.26} & 4.94 & 28.51 & \underline{25.66} & \multicolumn{1}{l|}{31.79} & \underline{27.46} \\
\multicolumn{1}{l|}{CoT} & 15.26 & 23.76 & \underline{31.36} & 28.10 & \multicolumn{1}{l|}{\underline{36.87}} & 16.90 & \underline{26.73} & 18.42 & 21.24 & \multicolumn{1}{l|}{\underline{27.59}} & 27.70 & 13.86 & 33.99 & 37.61 & \multicolumn{1}{l|}{\underline{44.15}} & 11.32 & 10.96 & \underline{29.39} & 22.57 & \multicolumn{1}{l|}{32.01} & 25.49 \\
\multicolumn{1}{l|}{DCoT} & 26.76 & 15.84 & 27.41 & \underline{32.96} & \multicolumn{1}{l|}{35.76} & 7.28 & 11.88 & 6.58 & 12.39 & \multicolumn{1}{l|}{5.74} & 19.72 & 21.78 & 8.99 & 10.84 & \multicolumn{1}{l|}{11.04} & 8.73 & 8.84 & 17.11 & 18.36 & \multicolumn{1}{l|}{22.96} & 16.55 \\
\multicolumn{1}{l|}{VCoT} & 5.87 & \underline{24.75} & 27.41 & 25.88 & \multicolumn{1}{l|}{35.54} & 25.59 & 24.75 & 11.40 & 15.27 & \multicolumn{1}{l|}{25.17} & \underline{31.46} & \underline{22.77} & \underline{41.01} & \underline{40.93} & \multicolumn{1}{l|}{35.98} & 18.76 & \underline{15.26} & 24.12 & 23.67 & \multicolumn{1}{l|}{\underline{32.23}} & 25.39 \\ \hline
\multicolumn{22}{c}{\textit{LLaVA-NeXT-8B}}                                                                                                                                                                                                                                                             \\
\cdashline{1-22} 
\multicolumn{1}{l|}{Direct} & \underline{42.96} & \underline{26.73} & 23.03 & \underline{30.09} & \multicolumn{1}{l|}{\underline{40.18}} & 25.35 & \underline{41.58} & \underline{83.55} & \underline{51.55} & \multicolumn{1}{l|}{\underline{68.21}} & 42.49 & \underline{41.58} & \underline{31.80} & \underline{31.86} & \multicolumn{1}{l|}{\underline{33.11}} & \underline{43.40} & \underline{51.19} & \underline{51.32} & \underline{35.84} & \multicolumn{1}{l|}{\underline{50.55}} & \underline{42.32} \\
\multicolumn{1}{l|}{CoT} & 30.05 & 12.87 & 24.34 & 11.50 & \multicolumn{1}{l|}{20.53} & 20.66 & 39.60 & 28.07 & 12.83 & \multicolumn{1}{l|}{39.07} & \underline{43.43} & 12.87 & 15.13 & 15.27 & \multicolumn{1}{l|}{22.96} & 10.35 & 43.29 & 44.74 & 21.90 & \multicolumn{1}{l|}{29.58} & 24.95 \\
\multicolumn{1}{l|}{DCoT} & 26.06 & 12.87 & 21.71 & 12.39 & \multicolumn{1}{l|}{10.60} & 17.00 & 24.75 & 41.23 & 14.16 & \multicolumn{1}{l|}{23.62} & 29.58 & 0.99 & 16.01 & 12.83 & \multicolumn{1}{l|}{20.53} & 13.26 & 28.47 & 44.08 & 25.44 & \multicolumn{1}{l|}{25.39} & 21.05 \\
\multicolumn{1}{l|}{VCoT} & 35.92 & 17.82 & \underline{25.44} & 10.40 & \multicolumn{1}{l|}{26.27} & \underline{25.82} & 34.65 & 44.96 & 30.53 & \multicolumn{1}{l|}{36.20} & 31.22 & 5.94 & 17.98 & 18.36 & \multicolumn{1}{l|}{28.26} & 13.13 & 41.79 & 33.77 & 19.69 & \multicolumn{1}{l|}{27.37} & 26.28 \\  \hline
\multicolumn{22}{c}{\textit{LLaVA-NeXT-72B}}                                                                                                                                                                                                                                                            \\

\cdashline{1-22} 
\multicolumn{1}{l|}{Direct} & \underline{73.24} & \underline{34.65} & \underline{35.75} & 36.95 & \multicolumn{1}{l|}{\underline{33.33}} & \underline{36.62} & 41.58 & 84.65 & 33.85 & \multicolumn{1}{l|}{83.89} & \underline{50.47} & \underline{30.69} & 30.04 & 38.94 & \multicolumn{1}{l|}{\underline{35.76}} & \underline{58.21} & 63.49 & 63.38 & 49.56 & \multicolumn{1}{l|}{69.98} & \underline{49.25} \\
\multicolumn{1}{l|}{CoT} & 68.54 & 26.73 & 30.92 & \underline{41.37} & \multicolumn{1}{l|}{19.43} & 34.98 & 43.56 & 86.40 & \underline{44.91} & \multicolumn{1}{l|}{83.00} & 46.24 & 24.75 & \underline{36.18} & \underline{45.58} & \multicolumn{1}{l|}{29.80} & 52.20 & 56.25 & 67.98 & 53.32 & \multicolumn{1}{l|}{71.96} & 48.21 \\
\multicolumn{1}{l|}{DCoT} & 56.81 & 15.84 & 33.99 & 38.94 & \multicolumn{1}{l|}{32.45} & 27.23 & \underline{54.46} & \underline{87.06} & 41.81 & \multicolumn{1}{l|}{\underline{84.55}} & 44.37 & 24.75 & 29.17 & 32.96 & \multicolumn{1}{l|}{26.71} & 32.34 & \underline{65.75} & \underline{72.15} & \underline{56.19} & \multicolumn{1}{l|}{\underline{77.92}} & 46.77 \\
\multicolumn{1}{l|}{VCoT} & 64.55 & 27.72 & 33.55 & 38.94 & \multicolumn{1}{l|}{32.01} & 35.68 & 46.53 & 85.96 & 44.03 & \multicolumn{1}{l|}{82.34} & 44.84 & 23.76 & 26.75 & 35.84 & \multicolumn{1}{l|}{34.00} & 49.68 & 56.46 & 69.30 & 55.97 & \multicolumn{1}{l|}{72.85} & 48.04 \\ \hline 
\multicolumn{1}{l|}{AVG} & 52.24 & 32.15 & 44.30 & 49.65 & \multicolumn{1}{l|}{37.42} & 32.87 & 37.16 & 64.67 & 38.65 & \multicolumn{1}{l|}{68.28} & 51.21 & 33.38 & 43.11 & 49.27 & \multicolumn{1}{l|}{33.78} & 35.09 & 50.94 & 53.69 & 44.31 & \multicolumn{1}{l|}{58.53} & 45.53 \\ \hline
\multicolumn{1}{l|}{AVG}    & \multicolumn{5}{c|}{43.15}                                                     & \multicolumn{5}{c|}{48.33}                                      & \multicolumn{5}{c|}{42.15}                                 & \multicolumn{5}{c|}{48.51}               & 45.53 \\ \hline
\end{tabular}
}
\caption{Main experimental results of four visual operations across five visualization types in multiple-choice format. ``Geo'' and ``Func'' are abbreviations for geometry and function. ``AVG'' represents the average accuracy across different models, visualization types, and visual operations. \textbf{Bold} numbers indicate the best results among all models and prompting methods, while \underline{underlined} numbers denote the best accuracy for individual models across all prompting strategies.}
\vspace{-1em}
\label{tab:main experiment 1.}
\end{table}

\begin{table}[htbp]
\centering
\setlength{\abovecaptionskip}{0.05cm}
\setlength{\belowcaptionskip}{0cm}
\renewcommand{\arraystretch}{1.3}
\scalebox{0.52}{
\begin{tabular}{l|lllll|lllll|lllll|lllll|l}
\hline
               & \multicolumn{5}{c|}{\textbf{MCG}} & \multicolumn{5}{c|}{\textbf{MCE (Deletion})} & \multicolumn{5}{c|}{\textbf{MCE (Modification)}} & \multicolumn{5}{c|}{\textbf{MCE (Annotation)}} & AVG  \\ \hline
                     & Geo   & Func  & Bar   & Line  & Pie  & Geo    & Func  & Bar   & Line  & Pie   & Geo   & Func  & Bar   & Line  & Pie  & Geo   & Func  & Bar   & Line  & Pie  &      \\ \hline
GPT-4o &\textbf{3.18} &\textbf{2.59} & 8.29 & 8.34 & 8.12 &\textbf{4.78} & 4.27 & 9.57 & 7.28 &\textbf{9.32} &\textbf{5.42} &\textbf{4.44} & 2.59 & 3.41 & 2.44 & 4.35 &\textbf{5.87} & 9.12 &\textbf{8.37} & 7.99 &\textbf{6.08} \\
Qwen-VL-Max & 2.86 & 2.40 & 8.23 & 8.56 & 8.50 & 3.59 &\textbf{4.64} &\textbf{9.78} & 7.71 & 9.10 & 4.54 & 3.59 &\textbf{2.61} & 4.18 & 2.49 &\textbf{5.00} & 5.27 & 9.06 & 7.28 & 8.34 & 5.97 \\
Qwen2.5-VL-3B-Instruct & 2.19 & 1.88 & 6.71 & 8.05 &\textbf{8.80} & 1.18 & 3.63 & 3.30 & 2.38 & 6.42 & 2.45 & 2.70 & 1.80 & 1.60 & 1.81 & 3.09 & 4.17 & 5.27 & 5.18 & 3.65 & 3.83 \\
Qwen2.5-VL-72B-Instruct & 2.76 & 1.94 &\textbf{8.45} &\textbf{8.83} & 8.60 & 3.73 & 4.05 & 9.59 &\textbf{8.33} & 8.75 & 4.70 & 3.29 & 2.48 &\textbf{4.21} &\textbf{2.57} & 4.15 & 5.55 &\textbf{9.17} & 7.73 & 7.75 & 5.94 \\
Gemma3-4B-IT & 1.90 & 1.08 & 6.37 & 6.67 & 8.49 & 1.21 & 1.01 & 1.39 & 1.07 & 1.97 & 2.00 & 1.00 & 2.01 & 1.96 & 2.32 & 2.51 & 1.01 & 7.00 & 7.39 & 4.58 & 3.30 \\
Gemma3-27B-IT & 2.62 & 1.27 & 7.06 & 7.52 & 8.79 & 1.92 & 1.00 & 3.31 & 2.91 & 6.07 & 2.09 & 1.00 & 1.99 & 1.97 & 2.45 & 3.31 & 1.00 & 8.61 & 6.98 & 5.40 & 4.05 \\
LLaVA-NeXT-8B & 1.71 & 1.06 & 5.92 & 6.85 & 8.57 & 1.67 & 3.43 & 1.10 & 1.12 & 1.81 & 1.81 & 3.34 & 1.74 & 1.69 & 1.74 & 2.63 & 3.89 & 5.06 & 5.93 &\textbf{8.72} & 3.51 \\
LLaVA-NeXT-72B & 2.34 & 1.21 & 7.36 & 7.17 & 8.79 & 1.54 & 3.38 & 4.34 & 1.88 & 6.65 & 2.30 & 1.67 & 1.83 & 1.66 & 1.91 & 3.62 & 4.00 & 7.86 & 7.71 & 4.06 & 4.09 \\ \hline 
AVG & 2.44 & 1.68 & 7.30 & 7.75 & 8.58 & 2.45 & 3.18 & 5.30 & 4.09 & 6.26 & 3.16 & 2.63 & 2.13 & 2.58 & 2.22 & 3.58 & 3.85 & 7.64 & 7.07 & 6.31 & 4.51 \\ \hline
AVG                  & \multicolumn{5}{c|}{5.55}              & \multicolumn{5}{c|}{4.26}            & \multicolumn{5}{c|}{2.54}           & \multicolumn{5}{c|}{5.69}              & 4.59 \\ \hline
\end{tabular}
}
\caption{Main experimental results of four visual operations across five visualization types in generation format. \textbf{Bold} numbers indicate the best results among all models.}
\vspace{-2em}
\label{tab:main experiment 2.}
\end{table}

\noindent
\textbf{Task Difficulty:} The complexity varies significantly across different task types. In both settings, MCE(Modification) task emerges as the most challenging with an average accuracy of merely 42.15\% in multiple-choice questions and an average score of just 2.54 in free-form generation. MCE(Annotation), however, achieves the highest scores in both settings, indicating that while models excel at understanding and processing numerical annotations, they struggle considerably with tasks requiring detailed comprehension and manipulation of image modifications.

\noindent
\textbf{Visualization Challenges:} Among different visualization types, function plots present the most significant challenges, achieving only 38.41\% accuracy in multiple-choice setting and an average score of 2.83 in free-form generation. And the difficulty level of geometric problems follows accordingly. While bar and pie charts demonstrate better performance with average accuracies around 50\%, line graphs unexpectedly show lower multiple-choice accuracy at 45.47\% compared to other statistical charts, revealing specific difficulties in understanding and manipulating line graph.

\noindent
\textbf{Prompting Effectiveness:} Contrary to previous findings, Direct prompting outperforms other strategies with 43.33\% accuracy, while DCoT shows the lowest performance at 37.59\%. This unexpected result likely stems from the more complex spatial relationships inherent in mathematical visualizations compared to natural images, where imprecise descriptions can actually hinder performance. 
Notably, while CoT prompting does not consistently outperform Direct prompting, larger models demonstrate enhanced CoT capabilities more frequently, suggesting the potential emergence of reasoning abilities with increased model scale.

\subsection{In-Depth Analysis}
This section contains further analysis of the evaluation of visual operations.

\begin{wrapfigure}{l}{0.48\textwidth}
    \centering
    \begin{minipage}{\linewidth}
        \centering
        
        \includegraphics[width=\linewidth]{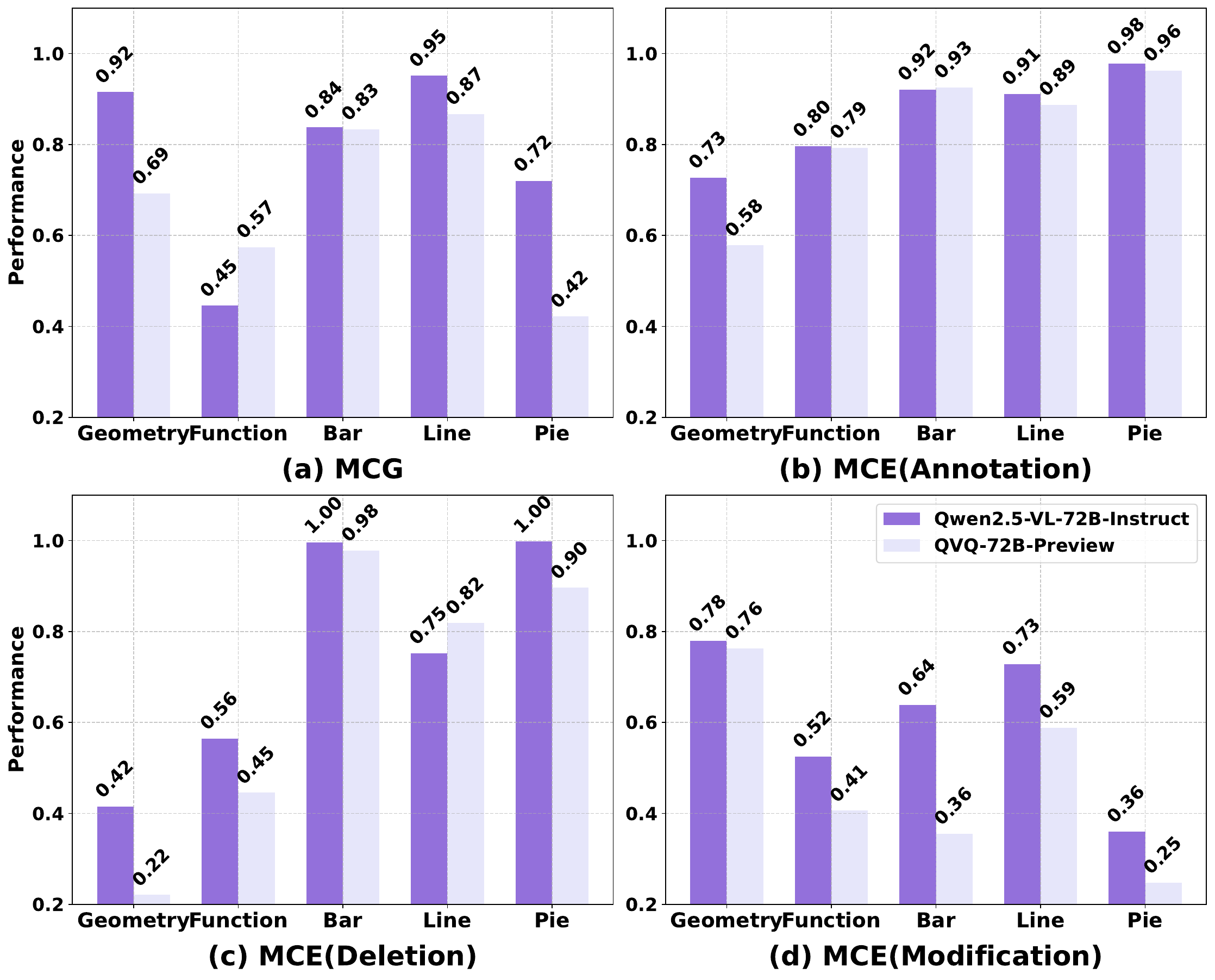}
        \captionof{figure}{Comparison between reasoning-enhanced and general-purpose models on multiple-choice format using Direct prompt.}
        \label{fig:qvq}
        \vspace{1em}
        \includegraphics[width=\linewidth]{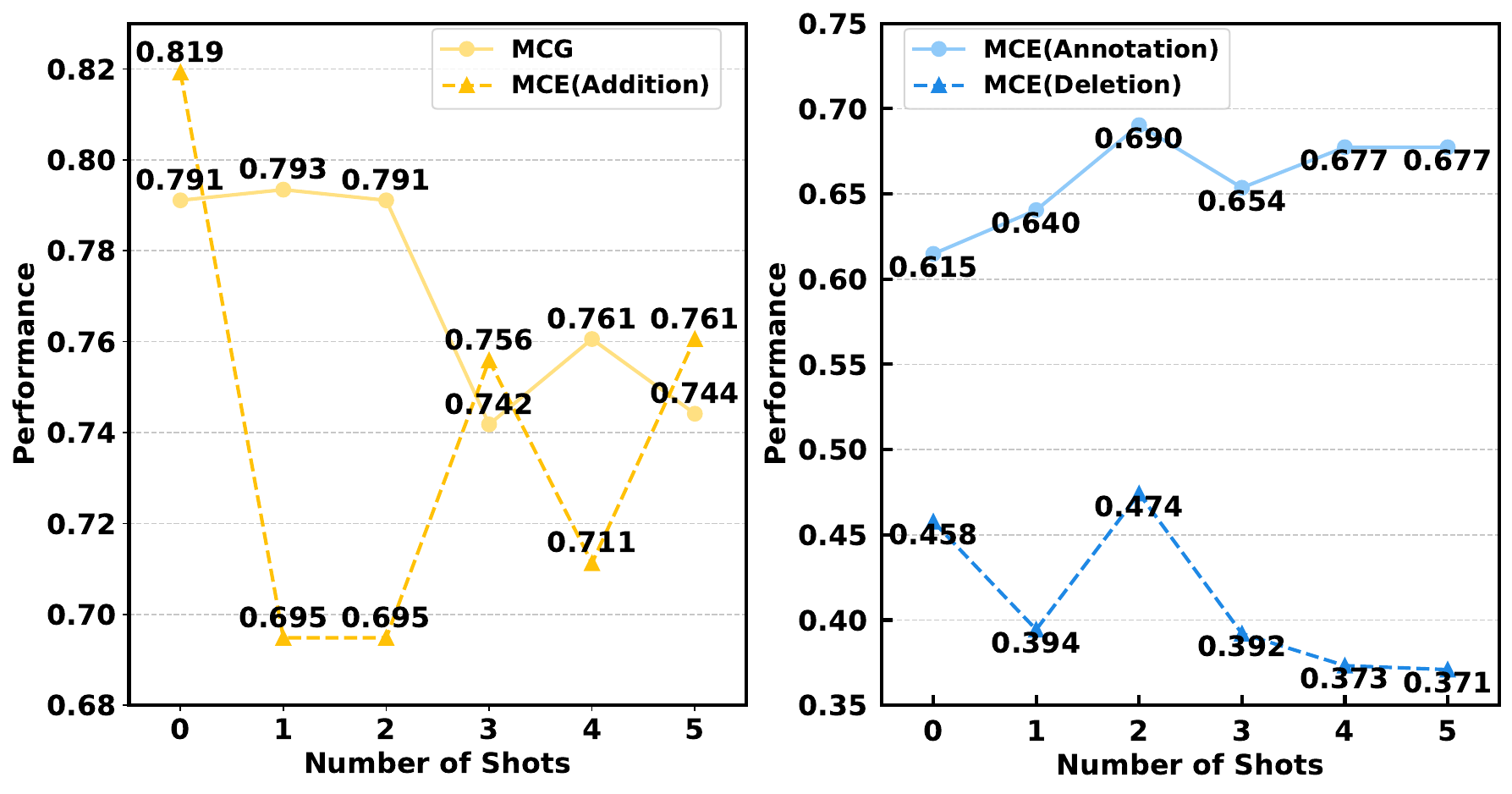}
        \captionof{figure}{Analysis of shot count effects on Qwen-VL-Max's multiple-choice performance across four visual operations using CoT prompt.}
        \label{fig:shot}
        \vspace{-3em}
    \end{minipage}
\end{wrapfigure}

\noindent
\textbf{Reasoning-Enhanced Models vs. General Models:} To investigate whether reasoning-enhanced models outperform general models of similar size in visual operations, we conduct a comparative analysis between QVQ-72B-Preview and Qwen2.5-VL-72B-Instruct. Figure \ref{fig:qvq} reveals an interesting pattern: the reasoning-enhanced model demonstrates superior performance in MCG of mathematical functions and MCE(Deletion) of line charts, but shows comparable or slightly inferior performance across other tasks. This performance disparity suggests that while reasoning enhancement effectively improves the model's reasoning capabilities, it does not necessarily translate to better performance in general visual operations.
Such observation raises important questions about the specialization of model capabilities and the trade-offs between focused reasoning enhancement and general visual understanding.

\noindent
\textbf{Impact of In-Context Learning:} We investigate the effectiveness of in-context learning (ICL) using Qwen-VL-Max on geometric figures. As demonstrated in Figure \ref{fig:shot}, ICL significantly enhances performance in MCE(Annotation) and MCE(Deletion) compared to zero-shot learning, which can be attributed to the relatively straightforward nature of these operations. However, when it comes to more complex tasks such as MCG and MCE(Modification), which involve diverse geometric elements (e.g., triangles, rectangles) and various line types (e.g., parallel, perpendicular, median lines and angle bisectors), ICL may introduce unintended noise. This complexity and the wide variability of possible operations can actually lead to performance degradation.


\noindent
\textbf{Shot Count Analysis:} Figure \ref{fig:shot} demonstrates that increasing the number of shots does not monotonically improve performance.  We observe that performance generally peaks at two shots and begins to decline with three or more. This highlights the importance of carefully tuning the shot count for optimal results. We attribute this performance degradation to two primary factors. On one hand, while initial examples help the model grasp the task, an excess of them can introduce noise or conflicting signals, especially with diverse inputs. On the other hand, models' finite context window inherently limits its ability to process numerous examples, leading to information loss and diminished performance.

\vspace{5pt}
\noindent
\textbf{Intersection Points vs. Extreme Points:} In Table \ref{table_func_gen} and \ref{table_func}, models generally perform best at identifying extreme points, followed by x-intercepts, with y-intercepts proving most challenging. Interestingly, we observe that Qwen2.5-3B-Instruct outperforms its 72B version on y-intercept tasks, and Gemma3-4B-IT surpasses its 27B version on all types. This counter-intuitive finding suggests that model size does not necessarily correlate with improved understanding of coordinate intersections.

\begin{table*}[t]
\centering

\begin{minipage}[t]{0.48\textwidth} 
    \centering
    \scalebox{0.7}{
        \begin{tabular}{l|lll|l}
        \hline
        Model                   & X    & Y    & Extreme & AVG  \\ \hline
        GPT-4o                  & 6.03 & 6.88 & 4.71    & 5.87 \\
        Qwen-VL-Max             & 5.26 & 6.50 & 4.06    & 5.27 \\
        Qwen2.5-VL-3B-Instruct  & 3.88 & 4.44 & 4.20    & 4.17 \\
        Qwen2.5-VL-72B-Instruct & 5.16 & 6.52 & 4.98    & 5.55 \\
        Gemma3-4B-IT            & 1.01 & 1.01 & 1.00    & 1.01 \\
        Gemma3-27B-IT           & 1.00 & 1.00 & 1.00    & 1.00 \\
        LLaVA-NeXT-8B           & 4.04 & 3.33 & 4.32    & 3.89 \\
        LLaVA-NeXT-72B          & 4.22 & 4.40 & 3.40    & 4.00 \\ \hline
        AVG                     & 3.82 & 4.26 & 3.46    & 3.85 \\ \hline
        %
        %
        \end{tabular}
    }
    \caption{Analysis of identifying function characteristics by MCE(Annotation) in free-form generation format.}
    \label{table_func_gen}
\end{minipage}
\hfill
\begin{minipage}[t]{0.48\textwidth} 
    \centering
    \scalebox{0.7}{
        \begin{tabular}{l|lll|l}
        \hline
        Model                & X     & Y     & Extreme & AVG   \\ \hline
        GPT-4o               & 74.65 & 60.15 & 82.57   & 72.45 \\
        Qwen-VL-Max          & 71.28 & 57.67 & 76.76   & 68.57 \\
        Qwen2.5-VL-3B-Instruct  & 69.68 & 64.11 & 68.66   & 67.48 \\
        Qwen2.5-VL-72B-Instruct & 74.47 & 53.47 & 92.43   & 73.45 \\
        Gemma3-4B-IT         & 10.28 & 14.60 & 16.73   & 13.87 \\
        Gemma3-27B-IT        & 2.84  & 12.38 & 14.79   & 10.00 \\
        LLaVA-NeXT-8B        & 54.79 & 33.91 & 34.86   & 41.19 \\
        LLaVA-NeXT-72B       & 65.43 & 60.40 & 55.63   & 60.49 \\ \hline
        AVG                  & 52.93 & 44.59 & 55.30   & 50.94 \\ \hline
        \end{tabular}
    }
    \caption{Analysis of identifying function characteristics by MCE(Annotation) in multiple-choice format.}
    \label{table_func}
    
\end{minipage}
\vspace{-3.5em}
\end{table*}

\noindent
\textbf{Forward vs. Backward Questions:} As shown in Table \ref{table_geo}, we compare performance on forward and backward questions in MCE(Annotation). The former identifies annotations based on geometric elements, e.g., what is the measure of angle ABC, while the latter identifies geometric elements based on annotations, e.g., which angle measures 120 degrees. We observe that models consistently perform better on backward questions. We hypothesize this asymmetry might stem from the fact that backward questions provide explicit numerical anchors that help constrain the search space, while forward questions require more comprehensive geometric understanding and spatial reasoning.
\begin{table*}[t]
\centering
\begin{minipage}[b]{0.48\textwidth}
    \centering
    \scalebox{0.75}{
        \begin{tabular}{l|ll|l}
        \hline
        Model                & Backward & Forward & AVG   \\ \hline
        GPT-4o               & 62.82    & 45.18   & 54.00 \\
        Qwen-VL-Max          & 72.48    & 49.60   & 61.04 \\
        Qwen2.5-VL-3B-Instruct  & 57.81    & 33.34   & 45.58 \\
        Qwen2.5-VL-72B-Instruct & 79.84    & 56.34   & 68.09 \\
        Gemma3-4B-IT         & 17.60    & 12.35   & 14.98 \\
        Gemma3-27B-IT        & 19.21    & 15.77   & 17.49 \\
        LLaVA-NeXT-8B        & 27.93    & 20.04   & 23.98 \\
        LLaVA-NeXT-72B       & 49.94    & 48.11   & 49.03 \\ \hline
        AVG                  & 48.45    & 35.09   & 41.77 \\ \hline
        \end{tabular}
    }
    \caption{Comparison of forward and backward problems in geometric MCE(Annotation) for multiple-choice format.}
    \label{table_geo}
\end{minipage}
\hfill
\begin{minipage}[b]{0.48\textwidth}
    \centering
    \scalebox{0.75}{
        \begin{tabular}{l|ll|l}
        \hline
        Model                   & Angle & Line & AVG  \\ \hline
        GPT-4o                  & 5.41  & 3.28 & 4.35 \\
        Qwen-VL-Max             & 5.19  & 4.81 & 5.00 \\
        Qwen2.5-VL-3B-Instruct  & 3.31  & 2.87 & 3.09 \\
        Qwen2.5-VL-72B-Instruct & 5.42  & 2.87 & 4.15 \\
        Gemma3-4B-IT            & 2.62  & 2.39 & 2.51 \\
        Gemma3-27B-IT           & 3.97  & 2.64 & 3.31 \\
        LLaVA-NeXT-8B           & 2.69  & 2.56 & 2.63 \\
        LLaVA-NeXT-72B          & 4.48  & 2.76 & 3.62 \\ \hline
        AVG                     & 4.14  & 3.02 & 3.58 \\ \hline
        \end{tabular}
    }
    \caption{Comparison of angle and line in geometric MCE(Annotation) for free-form generation.}
    \label{table_geo_gen}
\end{minipage}
\vspace{-2em}
\end{table*}

\noindent
\textbf{Points vs. Lines:} Regarding the generation setting for MCE(Annotation) in geometric problems, we further categorize the annotations into two types: angle markings and line labels. As illustrated in Table \ref{table_geo_gen}, angle annotations demonstrate notably higher scores compared to line annotations. We attribute this performance difference to the varying complexity of placement requirements: angle markings only need to be positioned near the identified vertex, whereas line labels require to be placed at the midpoint of the line segment. This more stringent positioning requirement for line labels contributes to the increased difficulty in their accurate placement.

\subsection{Error Analysis}
We conduct a manual analysis on 100 randomly selected outputs containing reasoning processes, equally divided between 50 correct and 50 incorrect processes. Our examination reveals concerning patterns in the models' performance: among the correct answers, only 33\% demonstrates both accurate results and valid reasoning processes, while 17\% achieves correct answers through wrong reasoning - indicating that correct outputs do not necessarily reflect sound problem-solving capabilities. We also identify four distinct categories of failures, with visual perception errors emerging as the predominant issue, accounting for 86\% of all errors. This remarkably high percentage highlights a crucial limitation in the models' ability to accurately process and interpret visual elements. Both instruction comprehension errors and output format violations each contribute to 6\% of the failures, while reasoning process errors constitute the remaining 2\%. This distribution of errors suggests that visual perception mechanisms require substantial improvement for reliable visual problem-solving.






\section{related work}
\textbf{Multimodal Mathematical Assessment:}
With the rapid development of MLLMs, numerous benchmarks have emerged to evaluate visual mathematical reasoning capabilities\citep{DBLP:conf/nips/WangPSLRZZL24,DBLP:conf/acl/HeLBHTSHHHZLQL024,DBLP:conf/cvpr/YueNZ0LZSJRSWYY24,DBLP:journals/corr/abs-2401-11944,DBLP:conf/nips/WangPSLRZZL24,DBLP:conf/emnlp/SunBQ0L24,DBLP:journals/corr/abs-2407-01284,DBLP:journals/corr/abs-2408-07543,DBLP:journals/corr/abs-2312-15915,DBLP:conf/acl/MasryLTJH22}. Previous work, such as Inter-gps \citep{DBLP:conf/acl/LuGJQHLZ20} and GeoQA \citep{DBLP:conf/acl/ChenTQLLXL21}, focuses on geometric problems, while others \citep{DBLP:conf/iclr/LuBX0LH0CG024,DBLP:conf/eccv/ZhangJZLGQZLCQGL24} cover broader tasks including function graphs and statistical charts. Though recent work \citep{DBLP:conf/nips/HuSFROZSK24,DBLP:journals/corr/abs-2412-12932,DBLP:journals/corr/abs-2411-05423} introduces a ``multi-modal input, multi-modal output'' paradigm, existing evaluations still assess only final answer accuracy, overlooking the quality of intermediate visual operations during reasoning.

\textbf{Visual Programming:}
The decomposition of complex visual tasks into manageable steps through programming interfaces has demonstrated significant potential in advancing visual reasoning capabilities\citep{DBLP:conf/iclr/YaoZYDSN023,DBLP:journals/corr/abs-2303-11381, DBLP:conf/iclr/ZengAICWWTPRSLV23,DBLP:conf/cvpr/HuSLVHLKF24}. Visprog \citep{DBLP:conf/cvpr/GuptaK23} and ViperGPT \citep{DBLP:conf/iccv/SurisMV23} showcase the effectiveness of using MLLMs to generate Python code for sequential visual operations. SKETCHPAD \citep{DBLP:conf/nips/HuSFROZSK24} further enables dynamic visual manipulations based on intermediate results, achieving particular success in geometric reasoning through visual operations. This progress inspires our work to move beyond evaluating just final mathematical solutions and toward assessing the quality and appropriateness of intermediate visual operations in the reasoning process.

\section{conclusion}
We present the first fine-grained benchmark for visual operations in mathematical reasoning, which systematically assesses MLLMs on four key operations across five visualization types. Our experiments reveal a significant performance gap between current models and humans, especially in function plots and MCE(Modification). We also find that the effectiveness of in-context learning is inconsistent and sensitive to the number of shots. We hope this work will guide the development of more capable models for visual mathematical reasoning.


\bibliography{iclr2026_conference}
\bibliographystyle{iclr2026_conference}

\appendix
\section{Prompt Design}
For different evaluation scenarios, we design specific prompts to guide the model's response. We consider two main question types: multiple-choice questions and free-form generation tasks.

For multiple-choice questions, the prompts are structured as follows:
\begin{itemize}
\item \textbf{Direct}:
\begin{verbatim}
Question: {question}
Choices:
{formatted_choices}
Please directly output the correct option.
\end{verbatim}
\item \textbf{CoT}:
\begin{verbatim}
Question: {question}
Choices:
{formatted_choices}
Let's think step by step. Please equip the correct 
option with \boxed{} at the end of your response.
\end{verbatim}

\item \textbf{DCoT}:
\begin{verbatim}
Question: {question}
Choices:
{formatted_choices}
Describe the image information relevant to the 
question. 
Please equip the correct option with \boxed{} at the 
end of your response.
\end{verbatim}

\item \textbf{VCoT}:
\begin{verbatim}
Question: {question}
Choices:
{formatted_choices}
Visualize the state after each reasoning step. 
Please equip the correct option with \boxed{} at the 
end of your response.
\end{verbatim}

\end{itemize}

For free-form generation tasks, the prompts follow similar patterns:
\begin{itemize}
\item \textbf{Direct}:
\begin{verbatim}
Question: {question}
Please directly output your code using markdown
format.
\end{verbatim}
\item \textbf{CoT}:
\begin{verbatim}
Question: {question}
Let's think step by step. Please wrap your code 
using markdown format at the end of your response.
\end{verbatim}
\item \textbf{DCoT}:
\begin{verbatim}
Question: {question}
Describe the image information relevant to the 
question. 
Please wrap your code using markdown format 
at the end of your response.
\end{verbatim}
\item \textbf{VCoT}:
\begin{verbatim}
Question: {question}
Visualize the state after each reasoning step. 
Please wrap your code using markdown format 
at the end of your response.
\end{verbatim}
\end{itemize}

Each prompt is designed to elicit specific types of responses while maintaining a consistent structure across different reasoning approaches.

\section{Human Annotation}

\subsection{Dataset Annotation Process}
The initial dataset annotation was conducted by authors of computer science Ph.D. candidates with extensive expertise in machine learning and computer vision. We possess comprehensive knowledge of both theoretical foundations and practical applications in the field, ensuring high-quality annotations.

\subsection{Main Experiment of Human Evaluation}
We carefully constructed a test subset comprising 400 samples, systematically selected from four visual tasks and five different categories (20 samples randomly selected from each combination). Three independent evaluators, all with strong computer science backgrounds but not involved in the initial annotation process, were recruited to assess these samples. These evaluators were specifically chosen for their programming expertise, as the evaluation involved code assessment components.

The evaluation process was structured as follows:
\begin{itemize}
\item Each evaluator independently assessed all 400 samples
\item Results were averaged across all three evaluators to minimize individual bias
\end{itemize}

\subsection{Evaluating LLM's Scoring Reliability}

To assess the reliability of LLMs in scoring tasks, we conducted a evaluation through manual annotation. Our analysis focused on examining whether the scores assigned by LLMs aligned with human judgment and established criteria. The evaluation process followed a structured approach with clearly defined assessment standards.

The following criteria were used to determine the reasonableness of LLM scoring:

\begin{itemize}
    \item \textbf{Consistency}: Whether the scores maintained consistency across similar responses
    \item \textbf{Justification}: Whether the LLM provided logical explanations for its scoring decisions
    \item \textbf{Alignment}: The degree of agreement with human expert scoring
    \item \textbf{Context Understanding}: The LLM's ability to consider contextual factors
\end{itemize}

We implemented a binary classification system for each scored instance:
\begin{itemize}
    \item \textbf{Reasonable}: The score aligns with evaluation criteria
    \item \textbf{Unreasonable}: The score shows significant deviation from expected standards
\end{itemize}

After thorough manual review of the scoring samples, we found that 92\% of the LLM's scoring decisions were deemed reasonable. This high percentage suggests that LLMs demonstrate strong capability in scoring.

\subsection{Error Analysis}
Our study involved analyzing 100 randomly selected outputs with reasoning processes, split evenly between 50 correct and 50 incorrect cases. The analysis uncovers concerning trends in model performance: of the correct answers, just 33\% show both accurate results and valid reasoning, while 17\% reach correct conclusions through flawed reasoning - suggesting that correct outputs may not truly indicate effective problem-solving abilities.

We identified four distinct error categories, with visual perception errors being the most significant, comprising 86\% of all errors. This notably high percentage indicates a fundamental weakness in how models process and interpret visual information. Instruction comprehension errors and output format violations each make up 6\% of the failures, with reasoning process errors accounting for the remaining 2\%. Based on this error distribution, it's clear that substantial improvements to visual perception mechanisms are necessary to achieve reliable visual problem-solving capabilities.

\section{Scoring Prompt}
\begin{figure*}[h]   
\centering
\setlength{\abovecaptionskip}{-0.10cm}
\setlength{\belowcaptionskip}{0cm}
\includegraphics[width=0.65\linewidth,scale=0.65]{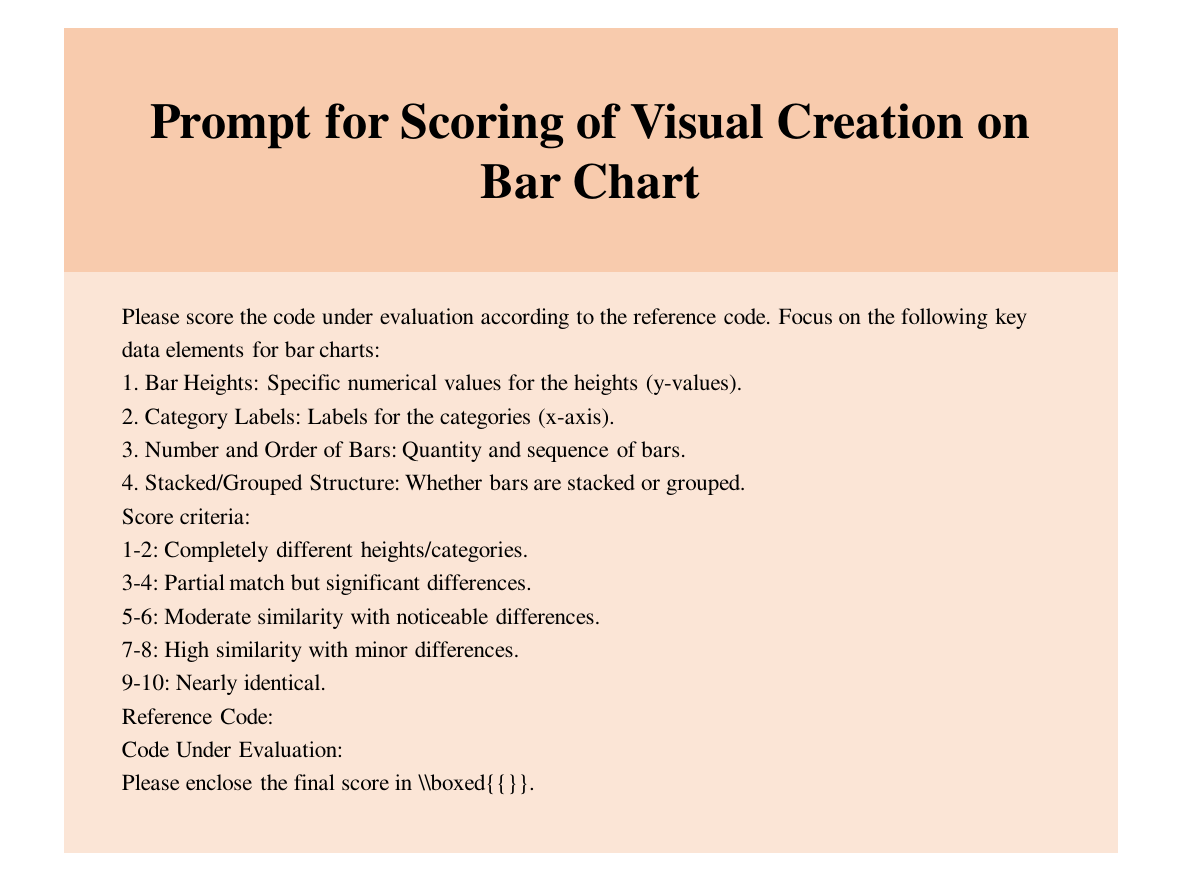}
\caption{Prompt for scoring of multi-modal code generation on bar charts. }
\label{task_1_bar}
\end{figure*}

\begin{figure*}[t]   
\centering
\setlength{\abovecaptionskip}{-0.10cm}
\setlength{\belowcaptionskip}{0cm}
\includegraphics[width=0.65\linewidth,scale=0.65]{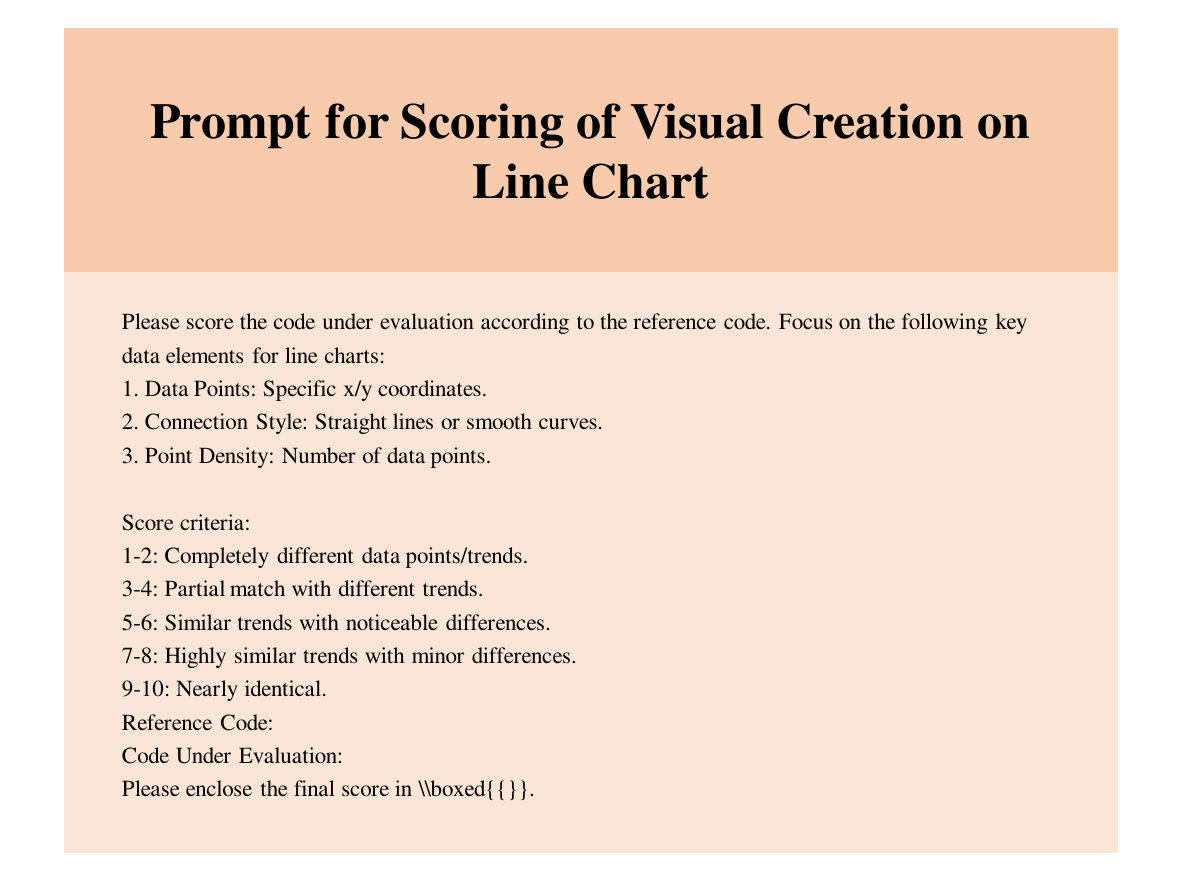}
\caption{Prompt for scoring of multi-modal code generation on line charts. }
\label{task_1_line}
\vspace{-10pt}
\end{figure*}

\begin{figure*}[t]   
\centering
\setlength{\abovecaptionskip}{-0.10cm}
\setlength{\belowcaptionskip}{0cm}
\includegraphics[width=0.65\linewidth,scale=0.65]{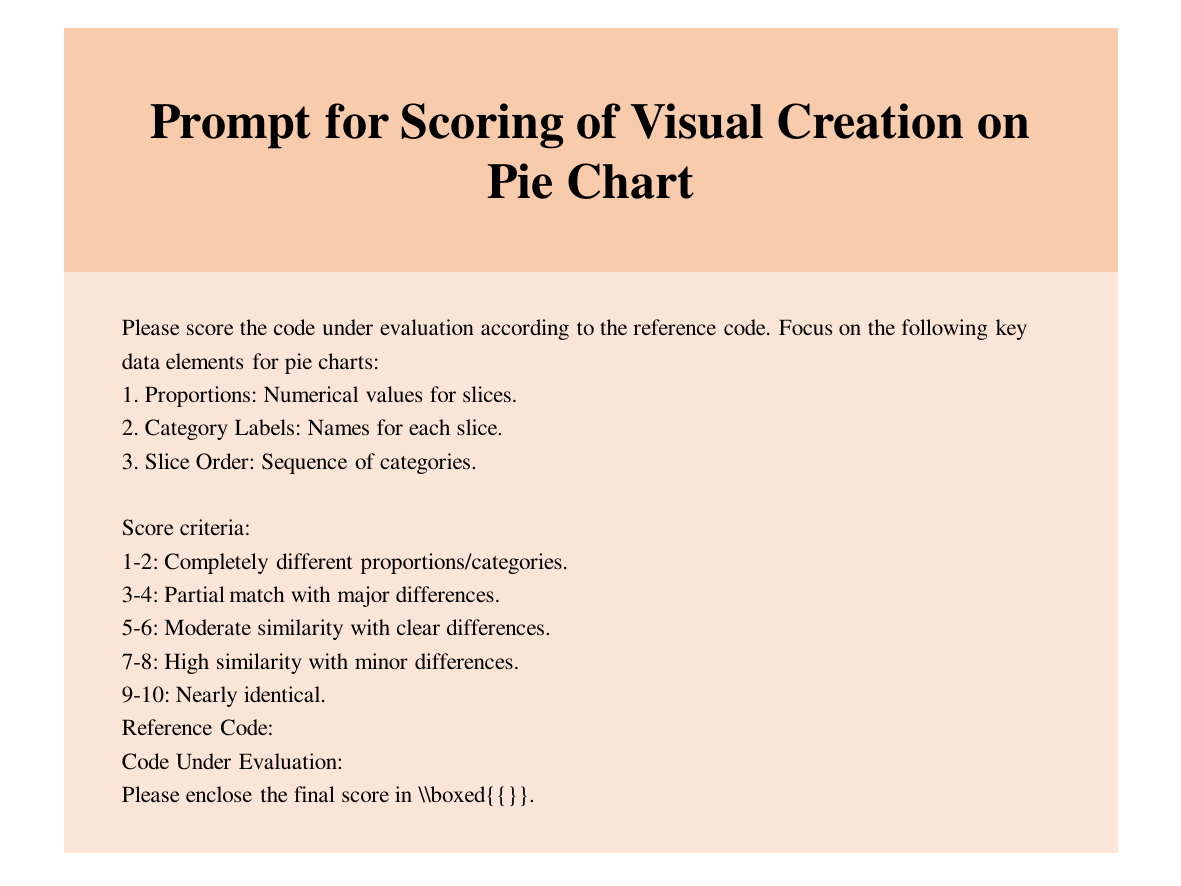}
\caption{Prompt for scoring of multi-modal code generation on pie charts. }
\label{task_1_pie}
\vspace{-10pt}
\end{figure*}

\begin{figure*}[t]   
\centering
\setlength{\abovecaptionskip}{-0.10cm}
\setlength{\belowcaptionskip}{0cm}
\includegraphics[width=0.65\linewidth,scale=0.65]{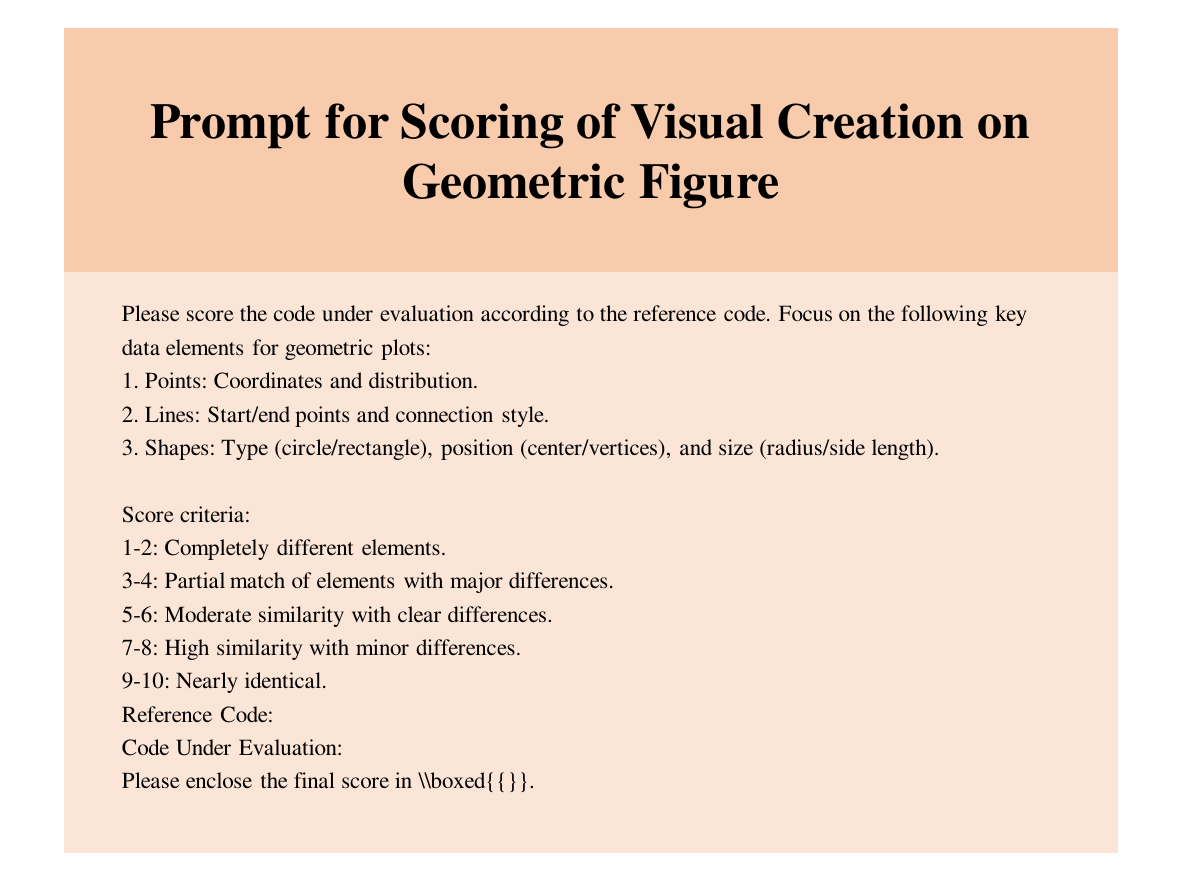}
\caption{Prompt for scoring of multi-modal code generation on geometric figures. }
\label{task_1_geo}
\vspace{-10pt}
\end{figure*}

\begin{figure*}[t]   
\centering
\setlength{\abovecaptionskip}{-0.10cm}
\setlength{\belowcaptionskip}{0cm}
\includegraphics[width=0.65\linewidth,scale=0.65]{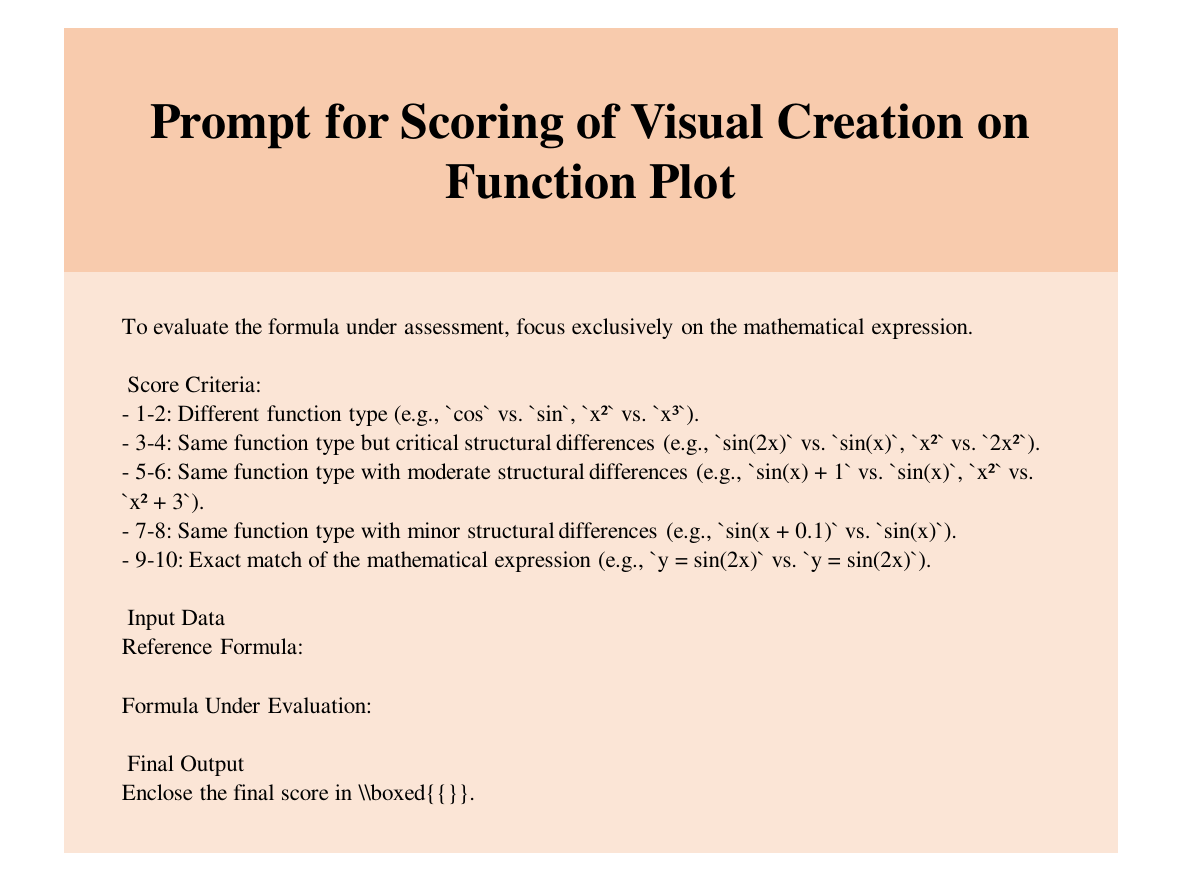}
\caption{Prompt for scoring of multi-modal code generation on function plots. }
\label{task_1_func}
\vspace{-10pt}
\end{figure*}

\begin{figure*}[t]   
\centering
\setlength{\abovecaptionskip}{-0.10cm}
\setlength{\belowcaptionskip}{0cm}
\includegraphics[width=0.65\linewidth,scale=0.65]{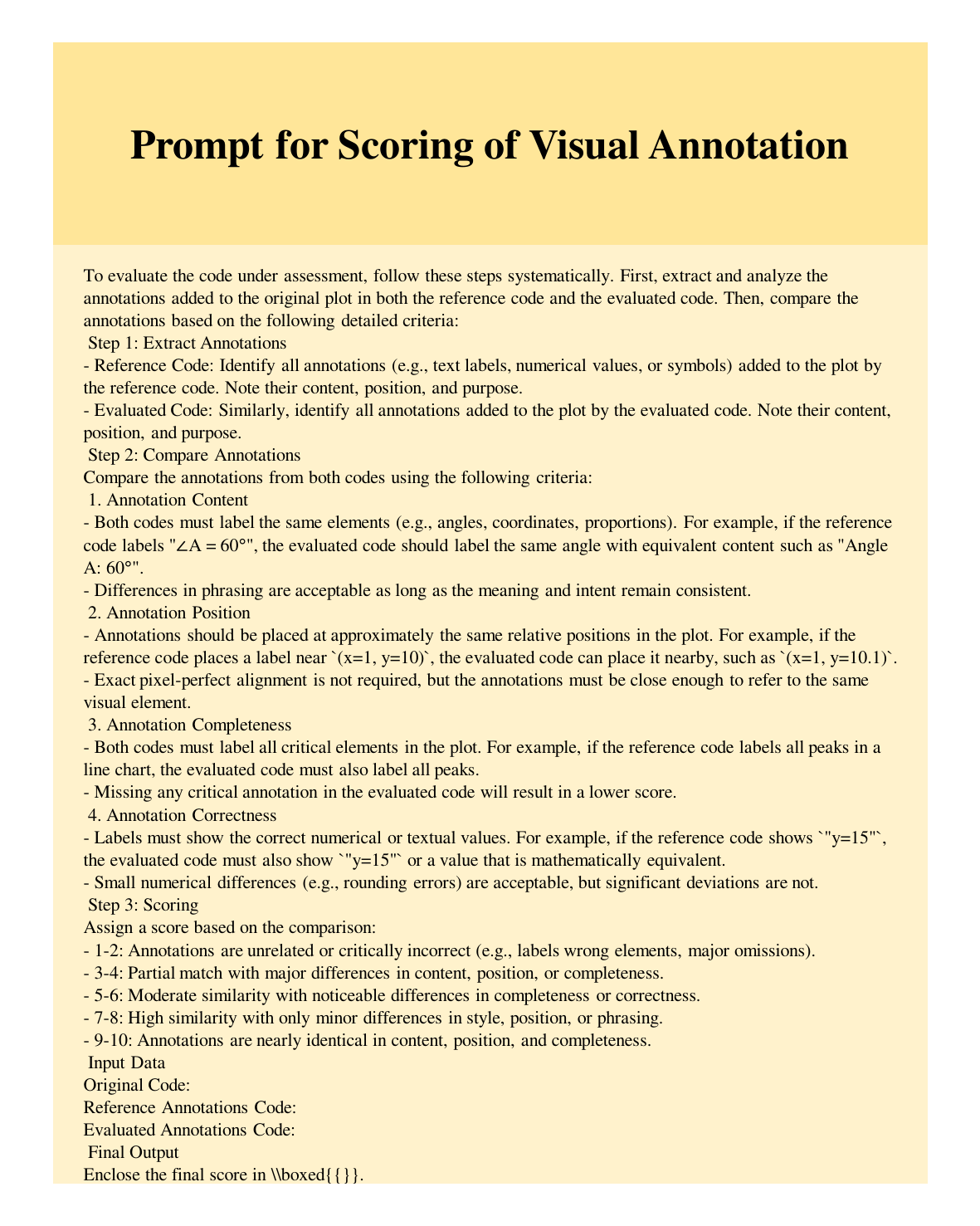}
\caption{Prompt for scoring of multi-modal code annotation. }
\label{task_2}
\vspace{-10pt}
\end{figure*}

\begin{figure*}[t]   
\centering
\setlength{\abovecaptionskip}{-0.10cm}
\setlength{\belowcaptionskip}{0cm}
\includegraphics[width=0.65\linewidth,scale=0.65]{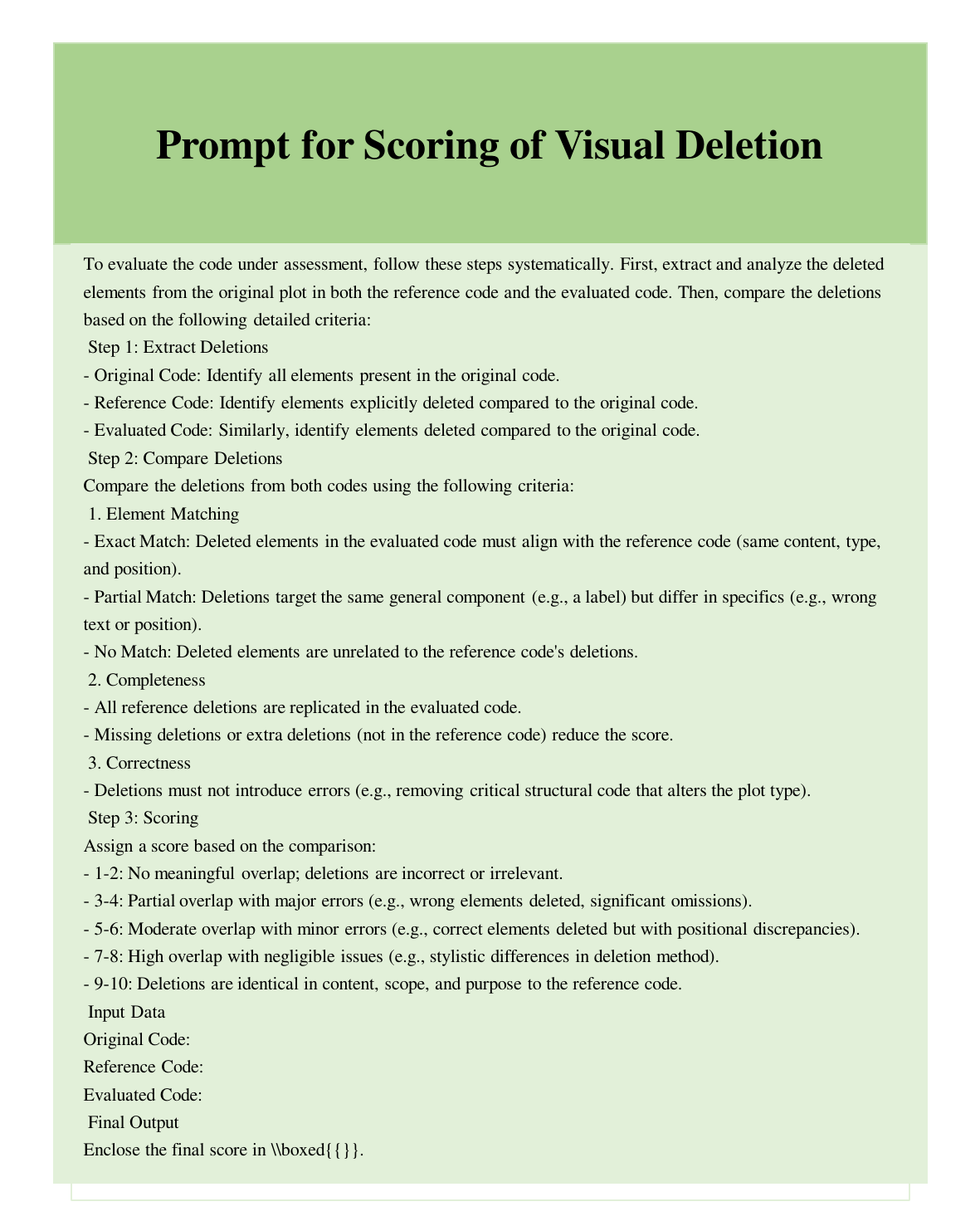}
\caption{Prompt for scoring of multi-modal code deletion. }
\label{task_3}
\vspace{-10pt}
\end{figure*}

\begin{figure*}[t]   
\centering
\setlength{\abovecaptionskip}{-0.10cm}
\setlength{\belowcaptionskip}{0cm}
\includegraphics[width=0.65\linewidth,scale=0.65]{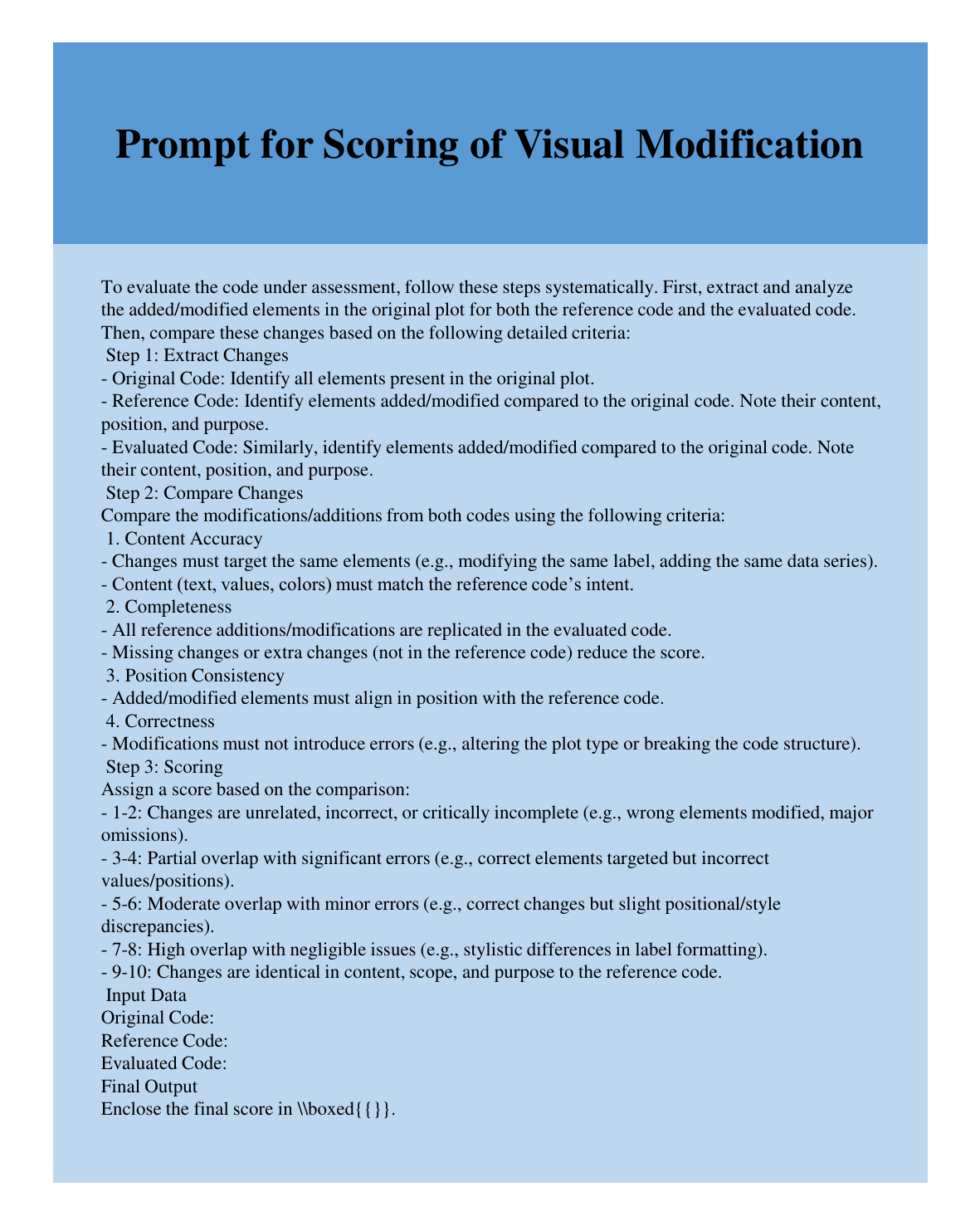}
\caption{Prompt for scoring of multi-modal code modification. }
\label{task_4}
\vspace{-10pt}
\end{figure*}

Figure \ref{task_1_bar}-\ref{task_4} represent the scoring prompts for different task types and different image types.
\begin{figure*}[t]   
\centering
\setlength{\abovecaptionskip}{-0.10cm}
\setlength{\belowcaptionskip}{0cm}
\includegraphics[width=\linewidth,scale=1.0]{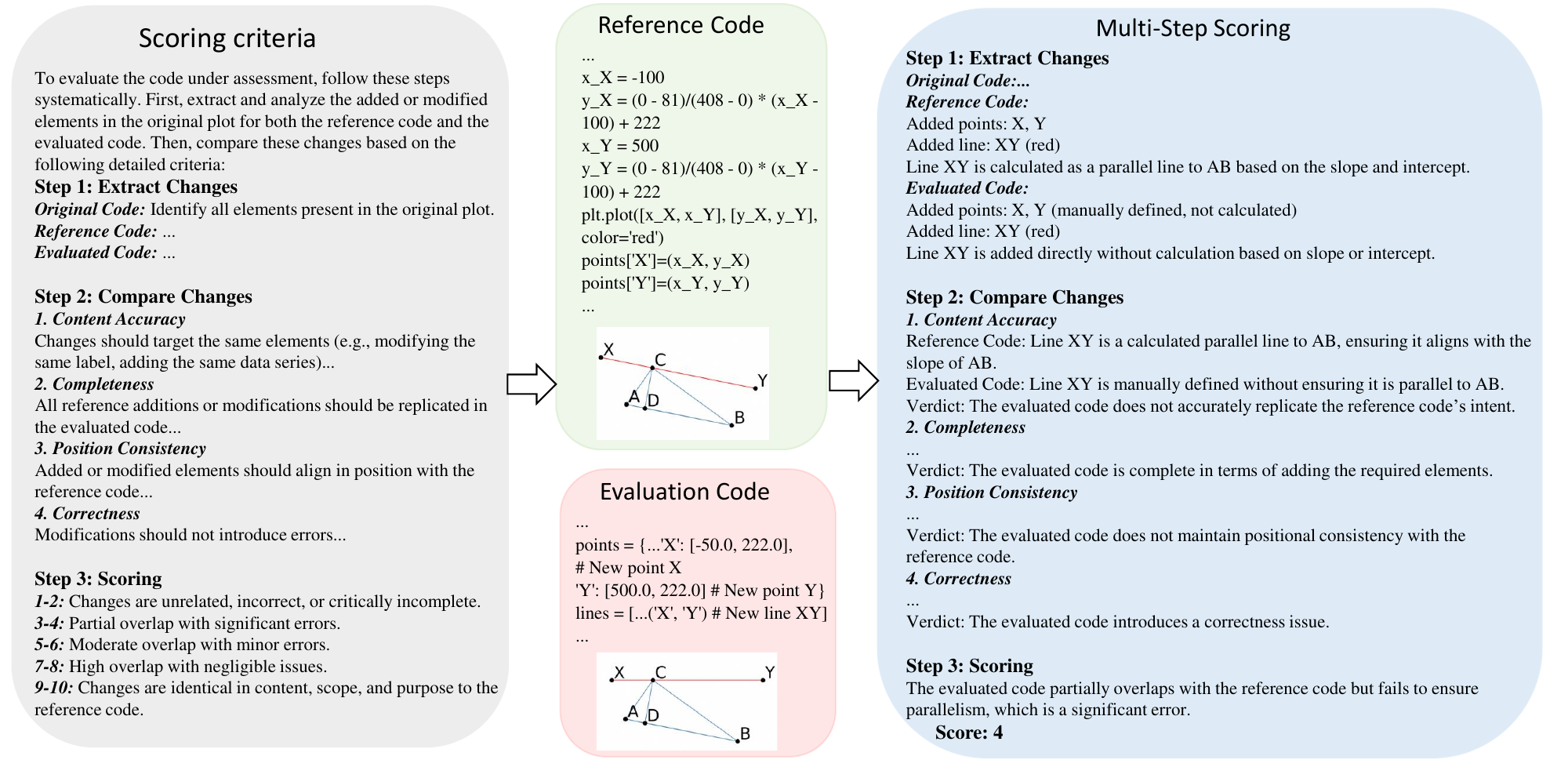}
\caption{Example of CoT-based evaluation strategy for MCE (Modification) in geometric figures. }
\label{cot}
\end{figure*}

\end{document}